\documentclass[letterpaper]{article} 
\usepackage[preprint]{aaai2027}
\usepackage[hyphens]{url}  
\usepackage{graphicx} 
\usepackage{natbib}  
\usepackage{caption} 
\usepackage{algorithm}
\usepackage[noend]{algorithmic}

\usepackage{newfloat}
\usepackage{amsfonts}
\usepackage{listings}
\DeclareCaptionStyle{ruled}{labelfont=normalfont,labelsep=colon,strut=off} 
\floatstyle{ruled}
\newfloat{listing}{tb}{lst}{}
\floatname{listing}{Listing}

\usepackage{booktabs}

\usepackage{microtype}
\usepackage{graphicx}
\usepackage{subfigure}
\usepackage{booktabs} 
\usepackage{algorithm}
\usepackage{algorithmic}
\usepackage{multirow}
\usepackage{xspace}
\usepackage{amsmath}
\usepackage{mathtools}
\usepackage{amsthm}
\usepackage[table]{xcolor}

\newtheorem{theorem}{Theorem}
\newtheorem{lemma}{Lemma}
\newtheorem{definition}{Definition}
\newtheorem{corollary}{Corollary}

\DeclareMathOperator*{\argmin}{arg\,min}

\newcommand{\ourmethod}{O$^2$CP\xspace}
\newcommand{\cyclone}{Cyclone\xspace}
\newcommand{\lane}{Lane\xspace}

\newcommand{\flu}{Flu hospitalization\xspace}
\newcommand{\weather}{Weather\xspace}
\newcommand{\elec}{Electricity\xspace}

\newcommand{\aci}{ACI\xspace}
\newcommand{\aciw}{ACI+\xspace}
\newcommand{\dtaci}{DtACI\xspace}
\newcommand{\dtaciw}{DtACI+\xspace}
\newcommand{\cpid}{CPID\xspace}
\newcommand{\cpidw}{CPID+\xspace}

\definecolor{ruipupurple}{RGB}{112,48,160}

\title{Optimization-Based Online Conformal Prediction for Multi-Step Forecasting}
\author{
    Ruipu Li,
    Daniel Menacho,
    Alexander Rodr{\'i}guez
}
\affiliations{
    University of Michigan\\
    liruipu@umich.edu, dmenordo@umich.edu, alrodri@umich.edu
}

\begin{document}

\maketitle

\begin{abstract}
Conformal prediction (CP) provides distribution-free coverage guarantees, making it well suited for uncertainty quantification in time series forecasting. However, existing methods often struggle with multi-step settings: they either calibrate horizons independently---ignoring temporal correlations---or enforce strict simultaneous coverage, resulting in overly conservative intervals. In this work, we propose O$^2$CP: Optimization-Based Online Conformal Prediction, a framework that augments a broad family of online CP methods with cross-horizon optimization while preserving their long-term coverage guarantees. We first characterize this family of methods, showing that long-term coverage is preserved as long as, at each forecast horizon, the selected control variable remains within an admissible set around the method's nominal output. Building on this result, O$^2$CP uses a two-layer design: the first layer constructs these admissible sets from the underlying online CP updates, and the second performs constrained optimization across horizons within them, jointly modeling the cross-horizon distributions to minimize a user-specified objective. Extensive experiments on real-world datasets---including autonomous driving, climate forecasting, and public health---demonstrate that O$^2$CP consistently outperforms state-of-the-art baselines, achieving target coverage with significantly sharper prediction intervals and reduced regret over long horizons.
\end{abstract}



\section{Introduction}
\begin{figure}[t]
\centering
\includegraphics[width=5.5 cm]{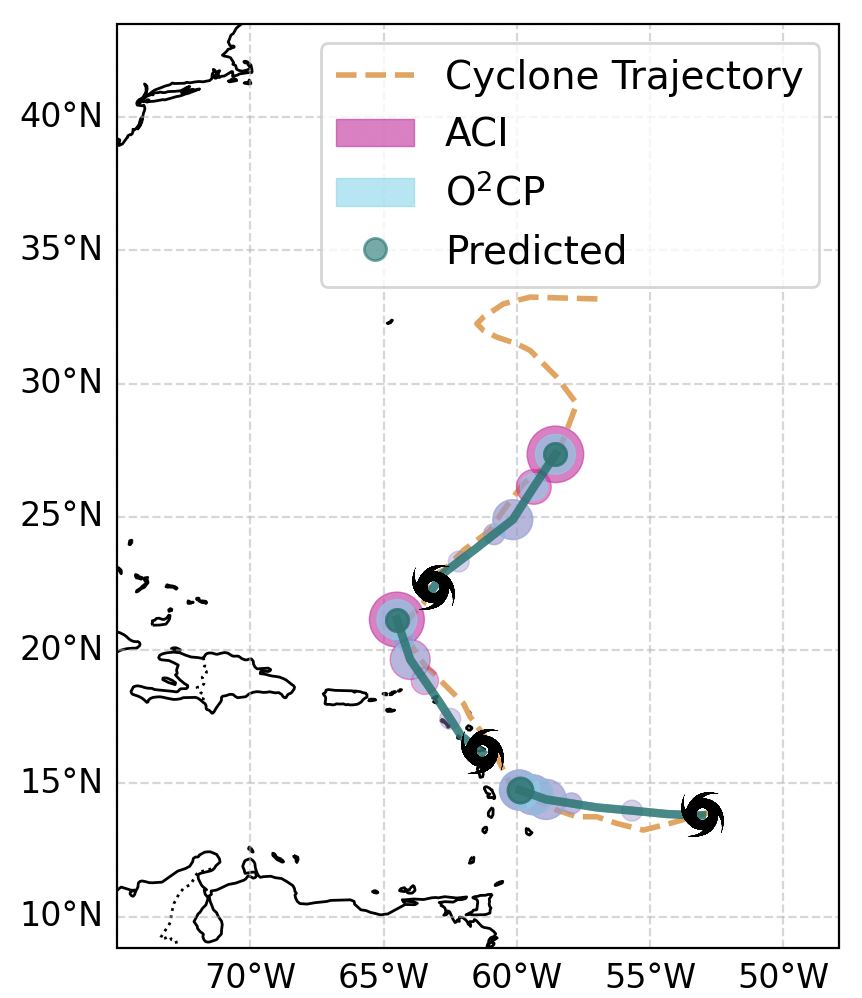}
\caption{\textmd{\cyclone trajectory forecasts with horizon-wise prediction sets. The orange dashed curve shows the ground truth, the green curve shows the prediction, and the shaded areas show the prediction sets at each horizon. Compared with \aci (pink), \ourmethod (sky blue) produces tighter sets.}
\label{fig:teaser}
}
\end{figure}

Multi-step time series forecasting is central to numerous real-world applications, including epidemic modeling, cyclone trajectory forecasting, and energy demand forecasting~\cite{mathis2024evaluation, price2025probabilistic, lam2023learning, CASCONE2023100360}. Online forecasting, in which observations arrive sequentially and models continually adapt to newly available data, closely reflects many real-world settings~\cite{pham2023learning}. Many applications require not only accurate predictions but also reliable uncertainty quantification for informed decision-making.

In this paper, we study uncertainty quantification for online multi-step time series forecasting through the lens of conformal prediction. 
Conformal prediction (CP) provides distribution-free, finite-sample coverage guarantees under exchangeability~\cite{vovk2005algorithmic}. Given a target miscoverage rate $\alpha$, CP constructs a prediction set $C$ that contains the ground truth with probability at least $1-\alpha$. Applying CP to online multi-step forecasting, however, presents two key challenges. First, temporal dependence and distribution shifts can violate exchangeability. Second, the predictive uncertainty (or equivalently, the forecast errors) across future time steps is usually correlated; for example, errors made at early horizons propagate forward and accumulate over longer horizons. Ignoring this structure and calibrating each horizon independently yields suboptimal prediction sets.

Recent work in online conformal prediction has studied the first challenge for single-step forecasting. 
A common idea in online conformal prediction is to replace a fixed miscoverage level with a time-varying \emph{control variable} \(z_t\), which determines the prediction set \(C_t(z_t)\). After each outcome is observed, \(z_t\) is updated according to whether the prediction set covered that outcome, allowing the prediction sets to adapt to distribution shifts while providing long-term coverage guarantees under suitable conditions~\cite{aci,cpid,gibbs2024aci,yang2024bellman}. These methods are primarily designed for single-step forecasting. A direct extension to multi-step forecasting maintains a separate control variable \(z_t^h\) for each horizon \(h\) and updates these variables independently. Consequently, it cannot exploit dependence across forecast horizons to improve the efficiency of the prediction sets.

A separate line of work explicitly models cross-horizon dependence in multi-step forecasting~\cite{cf-rnn,sun2023copula,lopes2024conforme}. These methods often target \emph{joint coverage}, requiring the prediction sets to cover the entire future trajectory simultaneously. This target can yield unnecessarily conservative prediction sets when only marginal coverage at each horizon is required, which is the case in epidemic forecasting~\cite{mathis2024evaluation}. Moreover, existing multi-step approaches are often designed for offline settings and assume exchangeable trajectories, which fail when distributions shift over time.

To address the two challenges identified above, we propose \textbf{O$^2$CP} (\textbf{O}ptimization-Based \textbf{O}nline \textbf{C}onformal \textbf{P}rediction), a framework that augments a broad family of online CP methods with cross-horizon optimization while preserving their long-term coverage guarantees (see Figure~\ref{fig:teaser}). We characterize this family and state the conditions required to preserve long-term coverage.
Our key insight is that the control variable used to construct a prediction set need not exactly equal the nominal value produced by the update rule. Instead, an admissible set can be constructed around each nominal value such that any sequence of values selected from these sets preserves long-term marginal coverage. O$^2$CP exploits this flexibility through a two-layer design: the first layer constructs an admissible set for each forecast horizon, and the second jointly optimizes the control variables within their respective admissible sets by modeling the cross-horizon distribution of forecast errors. In this way, the underlying online CP method continues to adapt to distribution shifts over time, while O$^2$CP separately accounts for dependencies across forecast horizons.

\subsection{Contributions}
\begin{enumerate}
    \item We introduce O$^2$CP, a framework that equips a broad family of online CP methods with cross-horizon optimization, and establish sufficient conditions for jointly optimizing control variables while preserving long-term coverage.

    \item We develop a lightweight sampling strategy that estimates the joint distribution of uncalibrated PIT values across horizons without multiple i.i.d.\ trajectories or large additional calibration sets.

    \item Experiments on five real-world datasets (autonomous driving, cyclone forecasting, public health, weather forecasting, and electricity demand) show that O$^2$CP significantly reduces long-horizon calibration regret and produces sharper prediction sets than state-of-the-art baselines.
\end{enumerate}

\section{Related Work}
\par\noindent\textbf{Conformal Prediction.}
Conformal prediction (CP) provides finite-sample, distribution-free marginal coverage under exchangeability~\cite{vovk2005algorithmic}. Split conformal prediction (SCP) provides an efficient implementation by dividing the data into a training set $\mathcal{D}_{\text{train}}$, used to fit the model $\hat{f}$, and a calibration set $\mathcal{D}_{\text{cal}}$. A nonconformity score measures the discrepancy between the observed label and its prediction, typically $s(y, \hat{y}) := |y - \hat{y}|$ in regression. Given a miscoverage rate $\alpha$, SCP computes the calibration scores and sets $\hat{q}$ to their $(1-\alpha)$-quantile. It then constructs
\[
\mathcal{C}(x_{\text{test}})
=
\left\{y : s\!\left(y,\hat{f}(x_{\text{test}})\right) \leq \hat{q}\right\},
\]
which contains the true label with probability at least $1-\alpha$.

\par\noindent\textbf{Online CP for Time Series.}
Distribution shifts and temporal dependence often make time series non-exchangeable, which can degrade standard CP. A common strategy in online conformal prediction is to maintain a time-varying control variable \(z_t\) that determines the prediction set \(C_t(z_t)\). The specific meaning of \(z_t\) depends on the method: in adaptive conformal inference (ACI)~\cite{aci}, \(z_t\) is the target miscoverage level \(\alpha_t\), whereas in conformal PID control (CPID)~\cite{cpid}, it is the complete conformity-score threshold \(q_t\). The control variable is updated online after each observation to adapt to distribution shifts.
For example, ACI updates the miscoverage level according to
\[
    \alpha_{t+1} = \alpha_t - \gamma (\mathrm{err}_t - \alpha),
\]
where \(\gamma\) is a step size and \(\mathrm{err}_t = \mathbf{1}(y_t \notin C_t)\) indicates a coverage violation at time \(t\). Prediction sets are constructed as
\[
    C_t = \bigl\{ y : s_t(y, \hat{y}_t) \leq \mathrm{Quantile}_{1-\alpha_t}(S_t) \bigr\},
\]
where \(s_t\) is a nonconformity score and \(S_t\) contains past calibration scores. This update achieves long-term coverage even under distribution shifts. Rather than adapting \(\alpha_t\), CPID updates the complete score threshold \(q_t\) directly and incorporates an error integrator and scorecaster to improve adaptation~\cite{cpid}. Bellman conformal inference (BCI) maintains a parameter \(\lambda_t\) that controls the tradeoff between coverage and prediction-interval width~\cite{yang2024bellman}. When a prediction interval misses the true outcome, the update shifts emphasis toward coverage and produces a wider interval. DtACI, an ensemble variant of ACI, provides a similar long-term coverage guarantee~\cite{gibbs2024aci}.

A direct multi-step extension maintains an independent control variable \(z_t^h\) for each forecast horizon $h$. Consequently, these methods neither model cross-horizon dependence nor optimize prediction sets jointly across horizons. O$^2$CP instead augments a broad family of online CP methods with cross-horizon optimization over their control variables while retaining their online adaptation toward long-term marginal coverage.


\par\noindent\textbf{Multi-step Time Series CP.}
Multi-step CP methods model the joint distribution across forecast horizons to improve prediction-set efficiency. Early approaches such as CF-RNN achieve joint coverage by applying a Bonferroni correction to the miscoverage rate, which often produces conservative prediction sets~\cite{cf-rnn}. Conforme improves efficiency by decomposing the joint distribution into a product of conditional distributions~\cite{lopes2024conforme}, while CopulaCP explicitly models cross-horizon dependence by learning an empirical copula from an additional calibration set~\cite{sun2023copula}. These methods target joint coverage---prediction sets that simultaneously cover the entire trajectory---and generally assume i.i.d.\ calibration trajectories. O$^2$CP instead targets long-term marginal coverage at each horizon in an online setting and estimates cross-horizon dependence from past observations without an additional i.i.d.\ trajectory calibration set.

Few CP methods target marginal coverage at each horizon in online multi-step time series forecasting. \citeauthor{acmcp}~\shortcite{acmcp} propose autocorrelated multi-step conformal prediction (acMCP), which augments CPID's quantile update with a structured error forecast that combines historical errors at the same horizon with errors at preceding forecast horizons. This formulation, however, makes general horizon-wide objectives, such as monotone interval widths, difficult to optimize. Moreover, adaptation to coverage errors and modeling of cross-horizon dependence occur within a single update, so the two goals must be balanced against each other. O$^2$CP separates these roles: the underlying online CP update drives time-wise adaptation toward long-term marginal coverage, while a constrained optimization captures cross-horizon dependence. This modular design preserves the base method's long-term coverage guarantees within admissible sets, and naturally accommodates general horizon-wide objectives.

\section{Problem Setup}
\label{sec:problem-setup}
Let $x_t \in \mathbb{R}^d$ denote the observed features at time $t$, where $d$ is the feature dimension, and let $y_t \in \mathcal Y$ denote the target. For any positive integer $n$, let $[n]:=\{1,\ldots,n\}$. For each forecast step $h \in [H]$, write $y_t^h := y_{t+h}$ for the outcome $h$ steps ahead of time $t$. For sequences, use the shorthand $x_{t_1:t_2} := (x_{t_1},\ldots,x_{t_2})$ and similarly for $y$. Given past observations $x_{1:t}$, a forecasting model $f$ produces the multi-step forecast
\[
\hat{y}_t^{1:H} = (\hat{y}_t^1,\ldots,\hat{y}_t^H) = f(x_{1:t}),
\]
where $H$ is the prediction horizon and $\hat{y}_t^h$ is the model's prediction of $y_t^h$. Given a user-specified coverage level $1-\alpha$, we construct prediction sets $C_t^{1:H} := (C_t^1,\ldots,C_t^H)$ at each time $t$ for all horizons $h \in [H]$.

\par\noindent\textbf{Obj.\ 1 (valid and efficient marginal coverage).}
Our primary objective is valid and efficient marginal coverage at each horizon. For every $h \in [H]$, the long-term miscoverage rate should match the target level $\alpha$:
\begin{equation}
\label{eq:long_term_obj}
\lim_{T \to \infty} \frac{1}{T}\sum_{t=1}^T \mathbf{1}\bigl(y_t^h \notin C_t^h\bigr) = \alpha,
\end{equation}
where $\mathbf{1}(\cdot)$ is the indicator function. We simultaneously seek small prediction sets, where $|C_t^h|$ denotes the size of the prediction set, giving the objective
\begin{equation}
\begin{aligned}
\label{eq:valid_efficient_obj}
\min_{C_1^{1:H},\ldots,C_T^{1:H}}
\; 
& \frac{1}{TH} \sum_{t=1}^T \sum_{h=1}^H |C_t^h|, \text{subject to } \eqref{eq:long_term_obj}.
\end{aligned}
\end{equation}

\par\noindent\textbf{Obj.\ 2 (horizon-wide user preferences).}
Users may also prefer particular behavior across the forecast horizon, such as prioritizing near-term forecasts or requiring smoothness. We represent these preferences through
\begin{equation}
\label{eq:horizon_wide_obj}
\min_{C_t^{1:H}} J(C_t^{1:H}), \quad t \in [1, T],
\end{equation}
where $J$ is a user-defined function that can incorporate auxiliary information or application-specific preferences into the construction of $C_t^{1:H}$.

\section{Method}
Achieving Obj.~1 and Obj.~2 requires coordinating two distinct optimization mechanisms. The first operates across time (typically through online updates based on observed coverage errors) to attain long-term coverage. The second operates across forecast horizons to model multi-step dependence and optimize user-specified horizon-wide objectives. \ourmethod realizes these mechanisms through a two-layer architecture that separates validity-preserving online adaptation from cross-horizon optimization.

In the first layer, we identify a common update rule shared by a family of online CP methods and generalize its point-valued output to an admissible set of control-variable values while preserving long-term coverage under the stated conditions. In the second layer, \ourmethod selects one value from each set by minimizing a horizon-wide objective. The first layer therefore preserves the coverage-feedback mechanism of the underlying method, while the second accounts for dependence across horizons. We describe the two layers in turn and summarize the complete procedure in Algorithm~\ref{alg:o2cp}.

\subsection{Generalized Online CP Updates and Admissible Sets}
\label{sec:online-update}

\paragraph{Common update rule.}
We identify a broad family of online CP methods whose coverage guarantees come from a common update rule. At time \(t\), a method in this family produces a control variable \(z_t\) and constructs a prediction set \(C_t(z_t)\).\footnote{Throughout this subsection, we fix a forecast horizon and suppress its index.} After observing \(y_t\), define the binary coverage error as
\(\mathrm{err}_t:=\mathbf 1\{y_t\notin C_t(z_t)\}\).
Although these methods use different control variables, their validity can be expressed through a state that accumulates coverage feedback. We call this quantity the \emph{coverage state} \(V_t\). It satisfies
\begin{equation}
\label{eq:common-update}
V_{t+1}-V_t=\gamma\bigl(\alpha-\mathrm{err}_t\bigr),
\end{equation}
where \(\gamma>0\) is a fixed step size. The control variable and coverage state have distinct roles: \(z_t\) constructs the prediction set, whereas \(V_t\) records coverage feedback.

Summing~\eqref{eq:common-update} gives
\[
\frac{1}{T}\sum_{t=1}^T\mathrm{err}_t
=\alpha+\frac{V_1-V_{T+1}}{\gamma T}.
\]
When \(V_{T+1}=o(T)\), the second term vanishes. The time-averaged miscoverage therefore converges to \(\alpha\).

Now we show how this recovers the update rule and coverage guarantee for a few popular online CP methods. For \aci, both quantities are \(\alpha_t\). The \aci update keeps \(\alpha_t\) bounded, so \(V_{T+1}=O(1)=o(T)\). For BCI, \(z_t=\alpha_t\) is the nominal miscoverage level selected by its Bellman optimization, while \(V_t=-\lambda_t\), where \(\lambda_t\) is the weight on its miscoverage penalty. The BCI update \(\lambda_{t+1}=\lambda_t-\gamma(\alpha-\mathrm{err}_t)\) therefore recovers~\eqref{eq:common-update}, and its safeguard keeps \(\lambda_t\) bounded, so \(V_{T+1}=O(1)=o(T)\)~\cite{yang2024bellman}. For \cpid, \(z_t=q_t\) is a score threshold, and the coverage state can be written as
\(V_t=-E_{t-1}\), where \(E_{t-1}:=\sum_{i=1}^{t-1}(\mathrm{err}_i-\alpha)\) accumulates the previous errors at the fixed forecast horizon. Thus, \(V_{t+1}-V_t=\alpha-\mathrm{err}_t\), recovering~\eqref{eq:common-update} with \(\gamma=1\). Under the bounded-score and saturation conditions of \cpid, its error integrator satisfies \(\lvert E_T\rvert\leq c h(T)+1\) for a sublinear function \(h\), and hence \(V_{T+1}=o(T)\). Many online CP methods also admit a coverage state satisfying~\eqref{eq:common-update} together with a method-specific sublinear-growth bound.

\paragraph{Generalized update rule.}
The update rule above leaves no freedom to coordinate calibration across horizons. We therefore construct a candidate set \(\mathcal Z_t\) around the nominal control-variable value \(z_t\) and use the generalized rule
\begin{equation}
\label{eq:generalized-update}
V_{t+1}-V_t=\gamma\bigl(\alpha-\mathrm{err}_t\bigr),
\qquad z_t^*\in\mathcal Z_t,
\end{equation}
where \(z_t^*\) is the control-variable value selected by the second layer. The final prediction set is \(C_t(z_t^*)\), and \(\mathrm{err}_t:=\mathbf 1\{y_t\notin C_t(z_t^*)\}\) is its realized coverage error.

\begin{definition}[Admissible sets]
\label{definition:admissible-sets}
A sequence of candidate sets \(\{\mathcal Z_t\}_{t\geq1}\) is \emph{admissible} for the generalized update rule if, for every sequence \(\{z_t^*\}_{t\geq1}\) satisfying \(z_t^*\in\mathcal Z_t\) for all \(t\), the induced coverage state satisfies \(V_{T+1}=o(T)\). We refer to each \(\mathcal Z_t\) as an admissible set.
\end{definition}

When \(\mathcal Z_t=\{z_t\}\), the generalized update rule recovers the original online CP method.

\begin{theorem}[Validity under the generalized update rule]
\label{theorem:generalized-update}
Let \(\{\mathcal Z_t\}_{t\geq1}\) be admissible for the generalized update rule. Then, for every sequence \(\{z_t^*\}_{t\geq1}\) satisfying \(z_t^*\in\mathcal Z_t\) for all \(t\),
\[
\frac{1}{T}\sum_{t=1}^T\mathrm{err}_t\longrightarrow\alpha
\quad\text{as }T\to\infty.
\]
\end{theorem}
\begin{proof}
Since \(V_{T+1}=o(T)\), substituting this condition into the sum of~\eqref{eq:generalized-update} gives the result.
\end{proof}

\paragraph{Admissible-set construction.}
We now construct an admissible set for \aci. The admissible sets for other methods used in this paper are in the appendix. For \aci, \(z_t=V_t=\alpha_t\), and~\eqref{eq:generalized-update} becomes
\(\alpha_{t+1}=\alpha_t+\gamma(\alpha-\mathrm{err}_t)\).
Given a relative radius \(\mu_t\in(0,1)\), when \(\alpha_t\in[0,1]\), define
\[
\mathcal Z_t^{\mathrm{ACI}}
:=\bigl\{a:\lvert a-\alpha_t\rvert
\leq\mu_t\min\{\alpha_t,1-\alpha_t\}\bigr\}.
\]
When \(\alpha_t\notin[0,1]\), let \(\mathcal Z_t^{\mathrm{ACI}}:=\{\alpha_t\}\). The radius \(\mu_t\) controls the relative freedom given to the second layer, while the factor \(\min\{\alpha_t,1-\alpha_t\}\) keeps the in-range candidate set inside \([0,1]\) and contracts it as \(\alpha_t\) approaches \(0\) or \(1\).

\begin{corollary}[Validity of the \aci admissible set]
\label{corollary:aci-admissible}
Assume \(\alpha\in(0,1)\), \(\alpha_1\in[0,1]\), a fixed \(\gamma>0\), \(C_t(a)=\mathcal Y\) for every \(a\leq0\), and \(C_t(a)=\emptyset\) for every \(a\geq1\). Then \(\{\mathcal Z_t^{\mathrm{ACI}}\}_{t\geq1}\) is admissible, and every sequence \(\{\alpha_t^*\}_{t\geq1}\) satisfying \(\alpha_t^*\in\mathcal Z_t^{\mathrm{ACI}}\) for all \(t\) guarantees long-term coverage~\eqref{eq:long_term_obj}.
\end{corollary}
The proof of Corollary~\ref{corollary:aci-admissible} is provided in the appendix. We also provide proofs for \cpid and \dtaci as representative methods. Other compatible methods can be proved similarly.

\begin{algorithm}[!t]
\caption{O2CP}
\label{alg:o2cp}
\textbf{Input:} Forecasting model \(f\), target miscoverage \(\alpha\), forecast horizon \(H\), and step size \(\gamma\). \\
\textbf{Initialize:} For each \(h\in[H]\), initialize \(z_1^h\) and \(V_1^h\), and set calibration scores \(S^h=\emptyset\).

\begin{algorithmic}[1]
\FOR{\(t = 1, 2, \ldots\)}
    \STATE \textbf{Receive} \(x_t\); compute \(\hat y_t^{1:H}=f(x_{1:t})\) and set \(c_t=\hat y_t^{1:H}\).
    \STATE \textit{\(\triangleright\) First layer: online updates and admissible sets}
    \FOR{\(h = 1, \ldots, H\)}
        \IF{\(t > h\)}
            \STATE Let \(\tau=t-h\), and observe \(y_t\) for the horizon-\(h\) prediction issued at \(\tau\).
            \STATE Compute its score and uncalibrated PIT \(\beta_\tau^h\), and update \(S^h\).
            \STATE Set \(\mathrm{err}_\tau^h\leftarrow\mathbf 1\{y_t\notin C_\tau^h(z_\tau^{h,*})\}\).
            \STATE Update \(V_t^h\leftarrow V_{t-1}^h+\gamma(\alpha-\mathrm{err}_\tau^h)\) and \(z_t^h\) according to the online CP method.
        \ENDIF
    \ENDFOR
    \STATE For each horizon without new feedback, carry its current state forward as \((V_t^h,z_t^h)\).

    \FOR{\(h = 1, \ldots, H\)}
        \STATE Construct an admissible set \(\mathcal Z_t^h\).
    \ENDFOR

    \STATE \textit{\(\triangleright\) Second layer: joint distribution estimation and cross-horizon optimization}
    \IF{\(t>H\)}
        \STATE Store \((c_{t-H},\Gamma_{t-H})\).
    \ENDIF
    \STATE Estimate \(\widehat{\mathbb P}_t\) based on \( \{(c_{\tau},\Gamma_{\tau})\}_{\tau=1}^{t-H} \).
    \[
        z_t^{1:H,*}
        \leftarrow
        \argmin_{u_1\in\mathcal Z_t^1,\ldots,u_H\in\mathcal Z_t^H}
        \mathbb E_{\beta^{1:H}\sim\widehat{\mathbb P}_t}[J].
    \]
    \FOR{\(h = 1, \ldots, H\)}
        \STATE Output the prediction set \(C_t^h(z_t^{h,*})\).
    \ENDFOR
\ENDFOR
\end{algorithmic}
\end{algorithm}

\subsection{Joint Distribution Estimation and Cross-Horizon Optimization}
\label{sec:horizon-wide-optimization}
Building on the result for admissible sets, the second layer models dependencies across forecast horizons while preserving the long-term coverage guarantee. It first estimates a joint distribution, based on which we then formulate the cross-horizon optimization problem.

\subsubsection{Joint distribution estimation}
The joint distribution of the binary coverage error encodes limited information, as this error only tells whether the selected prediction set covers the ground truth. Following Bellman conformal inference (BCI)~\cite{yang2024bellman}, we therefore introduce the \emph{uncalibrated probability inverse transform (PIT)}, which provides a more informative representation of the realized outcome. We then model the joint distribution of the PIT values across horizons.

Define \(g_t^h\) as the map from the method-specific control variable \(z_t^h\) to its nominal miscoverage level, \(\bar{\alpha}_t^h=g_t^h(z_t^h)\in[0,1]\). For \aci and \dtaci, \(g_t^h\) is the identity map. For \cpid, it converts the score threshold \(q_t^h\) to its empirical nominal miscoverage level, \(g_t^h(q_t^h)=1-\widehat F_t^h(q_t^h)\), where \(\widehat F_t^h\) is the empirical distribution function of the calibration scores.

For horizon \(h\), the uncalibrated PIT is defined as
\[
\beta_t^h:=\sup\{\bar{\alpha}\in[0,1]:s_t(y_t^h,\widehat y_t^h)\leq\operatorname{Quantile}_{1-\bar{\alpha}}(S^h)\}.
\]
Thus, \(\beta_t^h\) is the largest nominal miscoverage level at which the realized outcome \(y_t^h\) remains covered. A nominal miscoverage level no greater than \(\beta_t^h\) covers the outcome, whereas a level greater than \(\beta_t^h\) misses it. Then the realized coverage error for horizon \(h\) can be expressed as \(\operatorname{err}_t^h=\mathbf{1}\{g_t^h(z_t^{h,*})>\beta_t^h\}\). Intuitively, \(\beta_t^h\) quantifies how difficult horizon \(h\) is to cover at time \(t\): small values mean coverage fails even at low miscoverage levels, while large values mean \(y_t^h\) remains covered even at high miscoverage levels.

To model cross-horizon correlations, we estimate the joint distribution of \(\Gamma_t:=(\beta_t^1,\ldots,\beta_t^H)\). At each time \(t\), \(\Gamma_t\) is paired with a context vector \(c_t\) that summarizes the information available at that time. The context may be constructed from recent observations, or base-model predictions. In our implementation, we use the base-model's predictions $c_t := (\hat y_t^1,\ldots,\hat y_t^H)$.
The available historical pairs are therefore \(\{(c_\tau,\Gamma_\tau)\}_{\tau=1}^{t-H}\). We measure the similarity between \(c_t\) and each historical context \(c_\tau\) through \(d_t(\tau):=d(c_t,c_\tau)\), and assign normalized weights \(w_t(\tau)\propto k(d_t(\tau))\) using a chosen kernel or weighting function \(k(\cdot)\). We then approximate the joint distribution of \(\Gamma_t\) by the weighted empirical distribution
\[
\widehat{\mathbb P}_t(\Gamma)
:=\sum_{\tau=1}^{t-H}w_t(\tau)\,\delta_{\Gamma_\tau}(\Gamma).
\]

\subsubsection{Cross-horizon optimization}
Given \(\widehat{\mathbb{P}}_t\) and the admissible sets \(\mathcal Z_t^1,\ldots,\mathcal Z_t^H\), we select the control variables \(z_t^{1:H,*}\) by solving
\begin{equation}
\label{equation:multi-step-objective-J}
    \min_{u_1\in\mathcal Z_t^1,\ldots,u_H\in\mathcal Z_t^H}
    \;\mathbb{E}_{\beta^{1:H} \sim \widehat{\mathbb{P}}_t}
    \bigl[ J \bigr],
\end{equation}
where \(z_t^{1:H,*}=(u_1,\ldots,u_H)\). Our objective models cross-horizon dependence by treating the estimated joint distribution \(\widehat{\mathbb P}_t\) as the empirical ground truth. Its explicit form is
\begin{equation}
\begin{aligned}
    J =
    & \frac{1}{H}\sum_{h=1}^H \rho_{\alpha}\bigl(\beta^h - g_t^h(u_h)\bigr) + \\
    & \lambda \max\left\{\frac{1}{H} \sum_{h=1}^H \mathbf{1}\{ g_t^h(u_h) > \beta^h \}-\alpha, 0\right\},
\end{aligned}
\end{equation}
Here \(\rho_{\alpha}\) is the pinball loss at quantile level \(\alpha\). The first term encourages each \(g_t^h(u_h)\) to match the \(\alpha\)-quantile of \(\beta^h\). For the nested conformal families considered here, \(g_t^h(u_h)\leq g_t^h(u_h')\) implies \(C_t^h(u_h)\supseteq C_t^h(u_h')\), so a larger nominal miscoverage level produces a smaller prediction set. Targeting the calibrated \(\alpha\)-quantile therefore avoids unnecessarily low miscoverage levels and the wider prediction sets they induce, optimizing the overall efficiency of the prediction sets. The second term couples the horizons by penalizing sampled trajectories whose average miscoverage exceeds \(\alpha\), with \(\lambda\geq0\) controlling the penalty strength. The admissible-set constraints preserve the long-term coverage guarantee.

\section{Experiments}
\begin{table*}[t]
\centering
{\small
\setlength{\tabcolsep}{3.5pt}%
\begin{tabular}{@{}l|l|ccc|ccc|ccc@{}}
\toprule
Dataset & Metric & ACI+ & DtACI+ & CPID+ & ACI & DtACI & CPID & acMCP & CF-RNN & CopulaCPTS \\
\midrule
\multirow{4}{*}{lane} & SARegret@0.1 & \cellcolor{green!10}$0.0383^{\uparrow}$ & \cellcolor{green!10}$0.0673^{\uparrow}$ & \cellcolor{green!10}$\mathbf{0.0293}^{\uparrow}$ & $0.0491$ & $0.0730$ & $\underline{0.0317}$ & $0.0554$ & $0.0742$ & $0.0510$ \\
 & SARegret@All & \cellcolor{green!10}$0.0247^{\uparrow}$ & \cellcolor{green!10}$0.0314^{\uparrow}$ & \cellcolor{green!10}$\mathbf{0.0204}^{\uparrow}$ & $0.0275$ & $0.0346$ & $\underline{0.0214}$ & $0.0794$ & $0.2119$ & $0.0843$ \\
 & IntervalWidth & \cellcolor{green!10}$2.983^{\uparrow}$ & \cellcolor{green!10}$4.155^{\uparrow}$ & \cellcolor{green!10}$\mathbf{2.557}^{\uparrow}$ & $3.556$ & $4.508$ & $\underline{2.845}$ & $4.933$ & $9.424$ & $4.704$ \\
 & CS & $\underline{0.0823}$ & $0.1430$ & $0.0932$ & $\mathbf{0.0814}$ & $0.1388$ & $0.0872$ & $0.2189$ & $0.3489$ & $0.2938$ \\
\midrule
\multirow{4}{*}{cyclone} & SARegret@0.1 & \cellcolor{green!10}$0.0143^{\uparrow}$ & \cellcolor{green!10}$\underline{0.0132}^{\uparrow}$ & \cellcolor{green!10}$\mathbf{0.0115}^{\uparrow}$ & $0.0158$ & $0.0145$ & $0.0133$ & $0.0368$ & $0.0151$ & $0.0360$ \\
 & SARegret@All & \cellcolor{green!10}$0.0143^{\uparrow}$ & \cellcolor{green!10}$\underline{0.0125}^{\uparrow}$ & \cellcolor{green!10}$\mathbf{0.0117}^{\uparrow}$ & $0.0167$ & $0.0150$ & $0.0143$ & $0.0771$ & $0.1092$ & $0.1973$ \\
 & IntervalWidth & \cellcolor{green!10}$1.349^{\uparrow}$ & \cellcolor{green!10}$1.477^{\uparrow}$ & $\underline{1.322}$ & $1.377$ & $1.527$ & $\mathbf{1.294}$ & $2.791$ & $2.950$ & $5.174$ \\
 & CS & \cellcolor{green!10}$\underline{0.0681}^{\uparrow}$ & \cellcolor{green!10}$0.0805^{\uparrow}$ & \cellcolor{green!10}$\mathbf{0.0650}^{\uparrow}$ & $0.0797$ & $0.0991$ & $0.0703$ & $0.2028$ & $0.3424$ & $0.3927$ \\
\midrule
\multirow{4}{*}{elec} & SARegret@0.1 & \cellcolor{green!10}$0.0273^{\uparrow}$ & \cellcolor{green!10}$0.0226^{\uparrow}$ & \cellcolor{green!10}$0.0206^{\uparrow}$ & $0.0295$ & $0.0234$ & $0.0237$ & $\underline{0.0166}$ & $0.0239$ & $\mathbf{0.0131}$ \\
 & SARegret@All & \cellcolor{green!10}$0.0305^{\uparrow}$ & \cellcolor{green!10}$\mathbf{0.0175}^{\uparrow}$ & \cellcolor{green!10}$0.0234^{\uparrow}$ & $0.0345$ & $\underline{0.0183}$ & $0.0313$ & $0.0201$ & $0.0580$ & $0.0472$ \\
 & IntervalWidth & \cellcolor{green!10}$\mathbf{0.2359}^{\uparrow}$ & \cellcolor{green!10}$0.2691^{\uparrow}$ & \cellcolor{green!10}$0.2779^{\uparrow}$ & $\underline{0.2359}$ & $0.2724$ & $0.3024$ & $0.2394$ & $0.4758$ & $0.3655$ \\
 & CS & \cellcolor{green!10}$0.0570^{\uparrow}$ & \cellcolor{green!10}$\mathbf{0.0088}^{\uparrow}$ & \cellcolor{green!10}$0.0304^{\uparrow}$ & $0.0629$ & $0.0169$ & $0.0319$ & $\underline{0.0156}$ & $0.3538$ & $0.3182$ \\
\midrule
\multirow{4}{*}{hosp} & SARegret@0.1 & \cellcolor{green!10}$\mathbf{3171}^{\uparrow}$ & \cellcolor{green!10}$3755^{\uparrow}$ & \cellcolor{green!10}$5599^{\uparrow}$ & $3353$ & $4778$ & $5806$ & $5002$ & $\underline{3179}$ & $3723$ \\
 & SARegret@All & \cellcolor{green!10}$\mathbf{2546}^{\uparrow}$ & \cellcolor{green!10}$\underline{3088}^{\uparrow}$ & \cellcolor{green!10}$3198^{\uparrow}$ & $3165$ & $3671$ & $3560$ & $4767$ & $9064$ & $11449$ \\
 & IntervalWidth & \cellcolor{green!10}$\mathbf{9906}^{\uparrow}$ & \cellcolor{green!10}$16910^{\uparrow}$ & \cellcolor{green!10}$11824^{\uparrow}$ & $\underline{11033}$ & $18312$ & $13712$ & $16479$ & $36891$ & $29315$ \\
 & CS & \cellcolor{green!10}$0.0512^{\uparrow}$ & \cellcolor{green!10}$\mathbf{0.0150}^{\uparrow}$ & $0.0215$ & $0.0525$ & $0.0352$ & $0.0211$ & $\underline{0.0198}$ & $0.3584$ & $0.3111$ \\
\midrule
\multirow{4}{*}{weather} & SARegret@0.1 & \cellcolor{green!10}$2.681^{\uparrow}$ & \cellcolor{green!10}$2.296^{\uparrow}$ & \cellcolor{green!10}$2.219^{\uparrow}$ & $3.371$ & $2.568$ & $4.108$ & $\underline{2.150}$ & $\mathbf{2.137}$ & $3.479$ \\
 & SARegret@All & \cellcolor{green!10}$3.246^{\uparrow}$ & \cellcolor{green!10}$3.228^{\uparrow}$ & \cellcolor{green!10}$\mathbf{2.453}^{\uparrow}$ & $3.737$ & $3.420$ & $3.360$ & $\underline{2.878}$ & $4.698$ & $5.282$ \\
 & IntervalWidth & \cellcolor{green!10}$14.872^{\uparrow}$ & $17.952$ & \cellcolor{green!10}$\mathbf{14.529}^{\uparrow}$ & $15.023$ & $17.931$ & $\underline{14.572}$ & $14.731$ & $28.734$ & $20.714$ \\
 & CS & \cellcolor{green!10}$0.0887^{\uparrow}$ & $\underline{0.0474}$ & \cellcolor{green!10}$0.0804^{\uparrow}$ & $0.0951$ & $\mathbf{0.0460}$ & $0.0804$ & $0.0622$ & $0.3040$ & $0.2646$ \\
\bottomrule
\end{tabular}%
}
\caption{Mean performance across datasets and metrics. SARegret@0.1 is strongly adaptive regret at miscoverage rate $\alpha=0.1$; SARegret@All is averaged over all eleven miscoverage rates; and CS is calibration score. IntervalWidth is interval widths for one-dimensional datasets and prediction-set areas for two-dimensional datasets. The best value is bold, and the second-best is underlined. Pale-green shading and an up-arrow indicate that a ``$+$'' method outperforms its paired baseline. The corresponding standard deviations are reported in the appendix.}
\label{tab:kdd_main_results}
\end{table*}

\subsection{Experimental Setup}
\subsubsection{Datasets}
We evaluate O$^2$CP on five real-world datasets: autonomous navigation (\lane), cyclone trajectory forecasting (\cyclone), public health (\flu), weather forecasting (\weather), and electricity demand (\elec).
\textit{\lane} is derived from the Argoverse autonomous-vehicle dataset~\cite{Argoverse} and contains 250 sequences with a 5-step horizon. \textit{\cyclone} uses forecasts from \texttt{Weather Lab}\footnote{https://deepmind.google.com/science/weatherlab}~\cite{weatherlab3, weatherlab2}. It contains 26 cyclone scenarios, each with 5 predicted trajectories over an 8-step horizon. \textit{\flu} is constructed from CDC FluSight records on weekly influenza hospitalization\footnote{https://github.com/cdcepi/FluSight-forecast-hub}~\cite{mathis2024evaluation}. \textit{\weather}~\cite{lai2018modeling} and \textit{\elec}~\cite{xu2023sequential} are standard time-series benchmarks. The appendix provides further dataset details. 
We use the nonconformity score $s_t(y, \hat{y}) = |y-\hat{y}|$. For \lane and \cyclone, the forecaster produces multiple sampled trajectories; in this case, we compute $\hat{y}$ as the mean of the samples.

\subsubsection{Baselines}
We augment three online conformal prediction methods: \textit{\aci}~\cite{aci}, \textit{\dtaci}~\cite{gibbs2024aci}, and \textit{\cpid}~\cite{cpid}. We denote their augmented variants by \textit{\aciw}, \textit{\dtaciw}, and \textit{\cpidw}, respectively. We also compare these variants with acMCP~\cite{acmcp}, CF-RNN~\cite{cf-rnn}, and CopulaCPTS~\cite{sun2023copula}. acMCP is a CPID-based method for online multi-step time series that uses a scorecaster to model coverage dependence across horizons. CF-RNN and CopulaCPTS target joint coverage; we include them to compare the resulting prediction-interval widths. When available, we also evaluate the uncalibrated base forecasters. Our results show that their raw predictive samples are insufficiently calibrated, underscoring the need for conformal calibration; see the Appendix for details.

\subsubsection{Evaluation Metrics}
We evaluate \textit{coverage} and \textit{efficiency}, following common practice in conformal prediction~\cite{li2025neural}. Our primary metrics are \textit{calibration score} (CS), \textit{interval width}, and \textit{strongly adaptive regret} (SARegret); lower is better for all three.
Interval width measures prediction-set size: the distance between upper and lower bounds for one-dimensional data, or the area for two-dimensional data (\lane, \cyclone).
CS~\cite{kuleshov2018accurate} measures overall calibration quality by averaging the absolute gap between target and empirical coverage across a range of confidence levels; a perfectly calibrated model achieves $\text{CS}=0$.
SARegret~\cite{Zhangsariodr} captures worst-case calibration performance in nonstationary environments by measuring the maximum regret over all contiguous windows of a fixed length; a small value means the method tracks the best fixed predictor on every such window.
We evaluate each method at eleven miscoverage rates: 0.02, 0.05, 0.1, 0.2, 0.3, 0.4, 0.5, 0.6, 0.7, 0.8, and 0.9. For each dataset, we report the mean of each metric across multiple sequences, with reported standard deviations. The appendix provides formal definitions and further details.

\begin{figure*}[!t]
    \centering
    \includegraphics[width=1\linewidth]{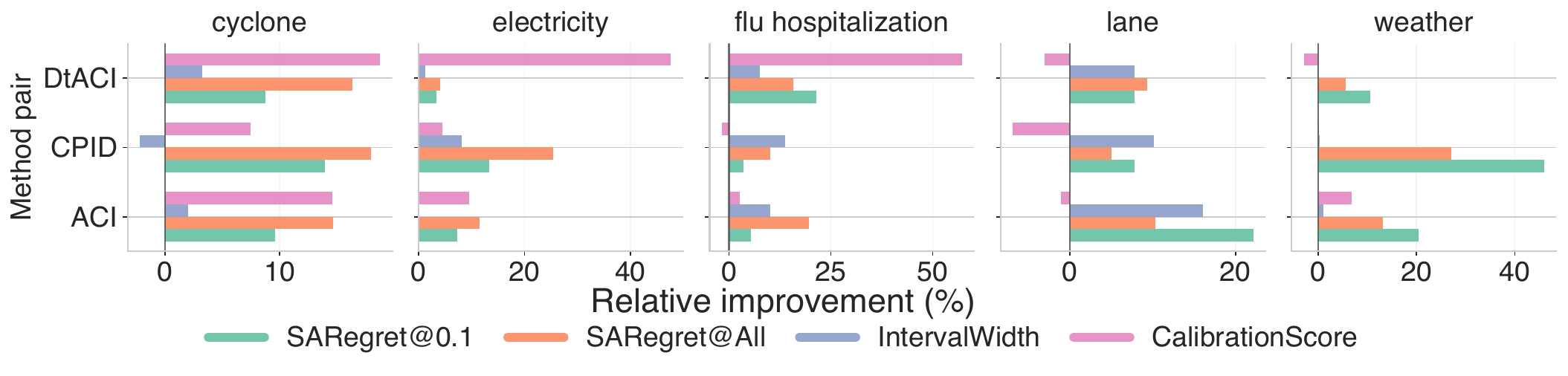}
    \caption{
        Relative improvement of our method (\textit{\aciw}, \textit{\dtaciw}, \textit{\cpidw}) over its paired baselines. Positive values indicate lower (better) metrics; gains are most consistent for SARegret and interval width.
    }
    \label{fig:rel_improv}
\end{figure*}

\subsection{Main Results}
\label{sec:main_results}
Table~\ref{tab:kdd_main_results} compares our method (\aciw, \dtaciw, \cpidw) with the online conformal prediction baselines (\aci, \dtaci, \cpid) and multi-step CP methods (acMCP, CF-RNN, and CopulaCPTS). Standard deviations within each dataset is reported in the appendix, which are small for \elec, \flu, and \weather, indicating consistent performance across their evaluation subsets. They are larger relative to the means for \lane and \cyclone, where uncertainty-estimation difficulty varies substantially across trajectories. For example, cyclone behavior differs across regions. We therefore conduct one-sided Wilcoxon signed-rank tests on these two datasets. The tests support many of the paired improvements; the appendix reports all $p$-values, including nonsignificant results.

\begin{figure*}[!htp]
    \centering
    \includegraphics[width=1\linewidth]{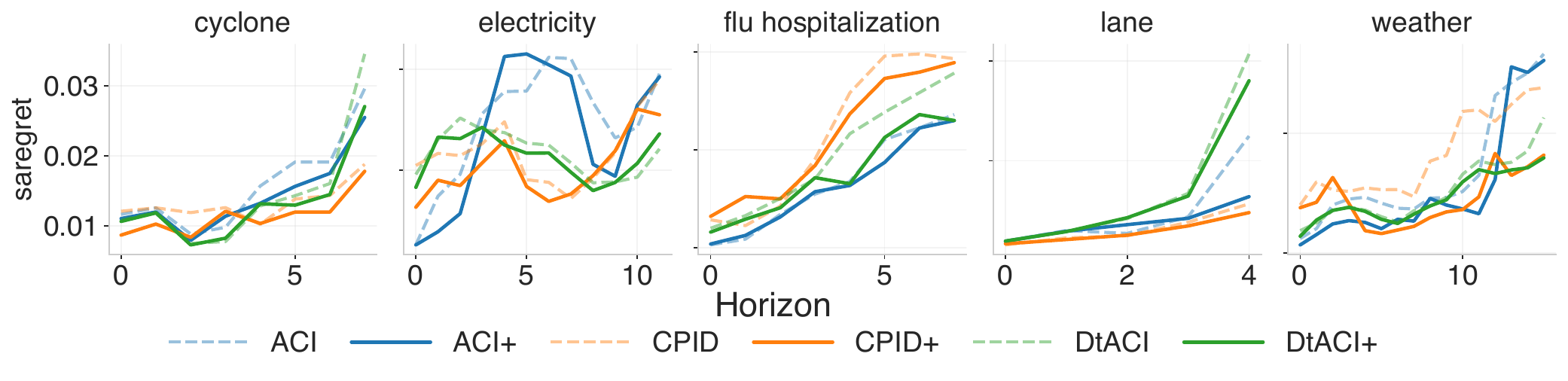}
    \caption{
        Horizon-wise SARegret at target miscoverage \(\alpha=0.1\). The ``+'' variants generally achieve lower regret than their paired baselines.
    }
    \label{fig:horizon_saregret}
\end{figure*}

\subsubsection{Improved Calibration Performance Over Baselines}
We first compare our method with the corresponding online conformal prediction baselines (\aci, \dtaci, \cpid). Pale-green shading and an up-arrow mark each ``+'' variant that outperforms its paired baseline. The ``+'' variants improve most paired metrics, with the clearest gains in SARegret and interval width. SARegret@All improves for every dataset and method pair, while SARegret@0.1 and interval width improve in nearly all comparisons. Calibration-score gains reach 57.2\% for \dtaciw on \flu and 47.6\% on \elec, but are mixed overall; for example, \dtaciw slightly increases interval width and CS on \weather. Even in these cases, SARegret@All improves. Figure~\ref{fig:rel_improv} summarizes these paired improvements. Overall, constrained cross-horizon optimization improves the efficiency by exploiting multi-step structure while preserving the long-term coverage objective. Figure~\ref{fig:horizon_saregret} shows that regret generally increases with the forecast horizon. The ``+'' variants generally achieve lower regret than their paired baselines, with gaps often widening at longer horizons. This suggests that cross-horizon optimization improves long-term calibration in most cases.

\subsubsection{Comparison with Multi-step Calibration Methods}
Table~\ref{tab:kdd_main_results} also compares our method with acMCP, CF-RNN, and CopulaCPTS. Although acMCP targets the same marginal coverage objective, the O$^2$CP variants outperform it on most datasets and metrics. The largest gains appear on \cyclone and \lane, where acMCP produces prediction intervals that are approximately twice as wide as ours (e.g., 2.791 versus approximately 1.3 on \cyclone) and has a higher calibration score (0.2028 versus 0.0650 for the best O$^2$CP variant). CF-RNN and CopulaCPTS target joint rather than marginal coverage and consequently produce substantially wider prediction intervals across all datasets. On \cyclone, for example, their interval widths are 2.950 and 5.174, compared with 1.477 for the widest O$^2$CP variant; their calibration scores are approximately 0.34 and 0.39, compared with 0.0805.

\subsubsection{Ablation Study}
\label{subsec:ablation_study}
The comparisons in Table~\ref{tab:kdd_main_results} between each ``$+$'' variant and its \emph{paired} baseline (e.g., \aciw/\aci) isolate the contribution of cross-horizon optimization: the baselines use the midpoint of each admissible interval (without optimization), whereas the ``$+$'' variants optimize within it. To examine the role of the online CP updates, we freeze the coverage-driven update for each method. For \aciw and \dtaciw, this freezes $\alpha_t$; for \cpidw, it freezes $q_t$. The frozen-update variants perform similarly on these metrics, supporting the empirical effectiveness of cross-horizon optimization. Nevertheless, the coverage-driven online CP updates and admissible constraints jointly provide the long-term coverage guarantee, so both remain essential even when the scalar metrics are similar. The appendix provides further details.

\section{Conclusion}
We present O$^2$CP, a framework for multi-step time series forecasting that models dependence across forecast horizons while maintaining long-term marginal coverage. Our analysis characterizes a broad family of online conformal prediction methods whose point-valued updates can be generalized to admissible sets. O$^2$CP uses this result in a two-layer design: the first layer constructs admissible sets around the underlying online CP updates, and the second jointly optimizes the control variables across horizons within these sets. Under the stated conditions, this design captures cross-horizon dependence while preserving long-term coverage. A lightweight sampling strategy estimates the joint distribution of uncalibrated PIT values without large calibration sets or multiple i.i.d.\ trajectories. Experiments on five real-world datasets spanning autonomous navigation, cyclone forecasting, public health, weather forecasting, and electricity demand show that O$^2$CP improves calibration and efficiency over baselines, reducing calibration score by up to 57.2\% while narrowing prediction intervals and yielding clear long-horizon regret gains. These results support constrained cross-horizon optimization for adaptive and efficient uncertainty quantification.

\par\textbf{Future work:} A promising direction is coupling O$^2$CP with downstream decision-aware losses that trade off uncertainty, cost, and risk, moving conformal prediction toward decision-aligned calibration.

\section{Limitations and Ethical Considerations}
\paragraph{Limitations}
Our method has several limitations. The effectiveness of distance-based trajectory retrieval varies across datasets; more informative context representations or kernel designs may improve joint-distribution estimation, particularly for high-dimensional data. The current framework also lacks an adaptive mechanism for correcting errors in the empirical joint distribution.

\paragraph{Ethical Considerations}
Our experiments use public benchmark datasets and collect no new or identifiable individual-level data. O$^2$CP should complement, not replace, expert judgment; safety-critical deployments should include domain expertise and robustness checks.

\bibliography{reference}


\appendix

\section{More on Related Work}
\subsection{Uncertainty Quantification for Time Series}
Time series forecasts are uncertain because of stochastic variation in the data-generating process, limited observations, and forecasting-model error. Quantifying these sources of uncertainty is important in applications such as public health and financial forecasting.
Ensemble methods~\cite{lakshminarayanan2017simple, ding2023dynamic} aggregate predictions from multiple models to estimate uncertainty. Their main limitation is computational cost, since they require fitting or running several models.
Bayesian methods~\cite{mackay1992practical, liang2005bayesian, chandra2021bayesian} instead learn a distribution over model parameters and induce a predictive distribution over the target~\cite{jospin2022hands}. They can also become computationally expensive as the number of parameters grows, and their performance depends on the underlying distributional assumptions.
Other approaches to uncertainty quantification in time series forecasting include diffusion models~\cite{rasul2021autoregressive} and quantile regression~\cite{koenker2001quantile}.

\section{Theory}

\subsection{ACI}
\label{proof:aci}

Fix a horizon and suppress its index. At time \(t\), O2CP selects
\(\alpha_t^*\in\mathcal Z_t^{\mathrm{ACI}}\), outputs
\(C_t(\alpha_t^*)\), and uses the realized error
\[
\mathrm{err}_t
:=
\mathbf 1\{y_t\notin C_t(\alpha_t^*)\}
\]
in the ACI update
\[
\alpha_{t+1}
=
\alpha_t+\gamma(\alpha-\mathrm{err}_t).
\]
By construction, the admissible set reduces to \(\{\alpha_t\}\) whenever
\(\alpha_t\notin[0,1]\).

\begin{lemma}[Boundedness of the ACI state]
\label{lemma:boundness-of-alpha}
\label{lemma: boundness of alpha}
Under the assumptions of Corollary~\ref{corollary:aci-admissible},
\[
-\gamma\leq\alpha_t\leq1+\gamma
\qquad\text{for every }t\geq1.
\]
\end{lemma}

\begin{proof}
Because \(\alpha\in(0,1)\) and \(\mathrm{err}_t\in\{0,1\}\),
\[
-\gamma
\leq
\alpha_{t+1}-\alpha_t
\leq
\gamma.
\]

For the upper bound, suppose that \(k+1\) is the first time at which
\(\alpha_{k+1}>1+\gamma\). Then \(\alpha_k\leq1+\gamma\), while the
increment bound implies
\[
\alpha_k\geq\alpha_{k+1}-\gamma>1.
\]
Hence \(\mathcal Z_k^{\mathrm{ACI}}=\{\alpha_k\}\) and
\(\alpha_k^*=\alpha_k>1\). The endpoint
convention gives \(C_k(\alpha_k^*)=\emptyset\), so
\(\mathrm{err}_k=1\). Consequently,
\[
\alpha_{k+1}
=
\alpha_k+\gamma(\alpha-1)
<
\alpha_k
\leq
1+\gamma,
\]
a contradiction.

For the lower bound, suppose that \(k+1\) is the first time at which
\(\alpha_{k+1}<-\gamma\). Then \(\alpha_k\geq-\gamma\), while the
increment bound implies
\[
\alpha_k\leq\alpha_{k+1}+\gamma<0.
\]
Thus \(\mathcal Z_k^{\mathrm{ACI}}=\{\alpha_k\}\) and
\(\alpha_k^*=\alpha_k<0\). The endpoint
convention gives \(C_k(\alpha_k^*)=\mathcal Y\), so
\(\mathrm{err}_k=0\). Therefore,
\[
\alpha_{k+1}
=
\alpha_k+\gamma\alpha
>
\alpha_k
\geq
-\gamma,
\]
again a contradiction. Since \(\alpha_1\in[0,1]\), both bounds hold
for every \(t\).
\end{proof}

\begin{proof}[Proof of Corollary~\ref{corollary:aci-admissible}]
Take \(V_t=\alpha_t\). The ACI update gives
\[
V_{t+1}-V_t
=
\alpha_{t+1}-\alpha_t
=
\gamma(\alpha-\mathrm{err}_t),
\]
for every selection
\(\alpha_t^*\in\mathcal Z_t^{\mathrm{ACI}}\).
Lemma~\ref{lemma:boundness-of-alpha} gives \(V_t=O(1)=o(t)\).
Thus \(\{\mathcal Z_t^{\mathrm{ACI}}\}_{t\geq1}\) is admissible, and
Theorem~\ref{theorem:generalized-update} gives the claimed long-term
coverage.
\end{proof}

\subsection{CPID}
For a fixed horizon, define the binary error of the final O2CP prediction
set and its accumulated centered value by
\[
\begin{aligned}
\mathrm{err}_t
&:=
\mathbf 1\{y_t\notin C_t(q_t^*)\},
&\qquad E_0&=0,\\
E_t
&:=
\sum_{i=1}^t(\mathrm{err}_i-\alpha).
\end{aligned}
\]
Write the CPID threshold before the O2CP selection as
\[
q_t
=
\widehat q_t+R_{t-1}(E_{t-1}),
\]
where \(\widehat q_t\) is the predictable scorecaster and \(R_{t-1}\)
is the CPID integrator. Assume
\(\widehat q_t\in[-b/2,b/2]\). Given a predictable relative radius
\(\mu_t^{\mathrm{CPID}}\in[0,1]\), define
\[
r_t^{\mathrm{CPID}}
:=
\mu_t^{\mathrm{CPID}}
\left(\frac b2-\lvert\widehat q_t\rvert\right)
\]
and the symmetric threshold set
\[
\mathcal Z_t^{\mathrm{CPID}}
:=
\left\{q:\lvert q-q_t\rvert\leq r_t^{\mathrm{CPID}}\right\}.
\]
If O2CP selects \(q_t^*\in\mathcal Z_t^{\mathrm{CPID}}\), write
\(d_t:=q_t^*-q_t\). The threshold used for prediction is then
\[
q_t^*
=
\widehat q_t+d_t+R_{t-1}(E_{t-1}).
\]
The parameter \(\mu_t^{\mathrm{CPID}}\) controls the freedom given to
the cross-horizon optimization: zero recovers the nominal CPID threshold,
whereas one gives the largest symmetric set around \(q_t\) that keeps the
adjusted scorecaster within the CPID bound.

\begin{corollary}[CPID admissible set]
\label{corollary:cpid-admissible}
Assume that the realized scores and predictable scorecasters lie in
\([-b/2,b/2]\), that
\(\mu_t^{\mathrm{CPID}}\in[0,1]\) is predictable, and that the integrator
\(R_t\) satisfies the CPID saturation condition. Then
\(\{\mathcal Z_t^{\mathrm{CPID}}\}_{t\geq1}\) is admissible for the
generalized update rule.
\end{corollary}

\begin{proof}
Take
\[
V_t=-E_{t-1},
\qquad
\gamma=1.
\]
Since \(E_t-E_{t-1}=\mathrm{err}_t-\alpha\),
\[
V_{t+1}-V_t
=
-E_t+E_{t-1}
=
\alpha-\mathrm{err}_t,
\]
which verifies the exact feedback identity for the final O2CP-selected
prediction set.

It remains to verify sublinear growth. With
\(\widehat q_t'=\widehat q_t+d_t\), the selected threshold becomes
\[
q_t^*
=
\widehat q_t'+R_{t-1}(E_{t-1}),
\]
which has exactly the CPID error-integration form with
\(\widehat q_t'\) acting as the scorecaster. By construction,
\[
\begin{aligned}
\lvert\widehat q_t'\rvert
&\leq \lvert\widehat q_t\rvert+\lvert d_t\rvert\\
&\leq \lvert\widehat q_t\rvert
+\mu_t^{\mathrm{CPID}}
\left(\frac b2-\lvert\widehat q_t\rvert\right)
\leq \frac b2.
\end{aligned}
\]
Since the realized score also lies in \([-b/2,b/2]\), its difference
from \(\widehat q_t'\) lies in \([-b,b]\). The CPID saturation result
therefore applies and gives, for its constants \(c>0\) and sublinear
function \(h\),
\[
|E_T|
\leq
c\,h(T)+1
=
o(T).
\]
Consequently, \(V_{T+1}=-E_T=o(T)\). The threshold sets are therefore
admissible, and Theorem~\ref{theorem:generalized-update} gives long-term
coverage.
\end{proof}

\subsection{DtACI}
\begin{algorithm}[tb]
\caption{O2CP adaptation of DtACI}
\label{alg:dtaci}
\textbf{Input:} Observed values \( \{\beta_t\}_{1\le t\le T} \), set of candidate \( \gamma \) values \( \{\gamma_i\}_{1\le i\le k} \), starting points \( \{\alpha_1^i\}_{1\le i\le k} \), and parameters \( \{\delta_t\}_{1\le t\le T} \), \( \sigma \), and \( \eta \). \\
\textbf{Initialize:} \( w_1^i \leftarrow 1 \), \( 1 \le i \le k \).
\begin{algorithmic}[1]
\FOR{\( t = 1, 2, \dots, T \)}
    
    \STATE Define probabilities \( p_t^i := w_t^i / \sum_{1 \le j \le k} w_t^j \), \( \forall 1 \le i \le k \).
    \STATE Select \( \alpha_t = \alpha_t^i \) with probability \( p_t^i \), and let \(i_t\) be the selected index.
    
    \STATE Choose \(\tilde{\alpha}_{t}\) from the selected expert's admissible set. If \(\alpha_{t} \notin [0,1]\), set \(\delta_{t}=0\) and \(\tilde{\alpha}_{t}=\alpha_{t}\).
    \STATE Construct the prediction set \(\hat{C}_{t}^{\tilde{\alpha}_{t}}\), with the monotonicity convention \(\hat{C}_t^{0}=\mathcal{Y}\) and \(\hat{C}_t^{1}=\emptyset\).
    \STATE \( \bar{w}_t^i \leftarrow w_t^i \exp(-\eta \ell(\beta_t, \alpha_t^i)) \), \( \forall 1 \le i \le k \).
    \STATE Compute the total intermediate weight:
    \[
        \bar{W}_t \leftarrow \sum_{1 \le i \le k} \bar{w}_t^i
    \]
    \STATE \( w_{t+1}^i \leftarrow (1-\sigma)\bar{w}_t^i + \bar{W}_t\sigma/k \).
    \STATE \( \text{err}_t := \mathbb{1}\{Y_t \notin \hat{C}_t(\tilde{\alpha}_{t})\} \).
    \STATE \( \text{err}_t^i := \mathbb{1}\{Y_t \notin \hat{C}_t(\alpha_t^i)\} \), \( \forall 1 \le i \le k \).
    \FOR{\( 1 \le i \le k \)}
        \IF{\( i = i_t \)}
            \STATE \( \alpha_{t+1}^i = \alpha_t^i + \gamma_i(\alpha - \text{err}_t) \).
        \ELSE
            \STATE \( \alpha_{t+1}^i = \alpha_t^i + \gamma_i(\alpha - \text{err}_t^i) \).
        \ENDIF
    \ENDFOR

\ENDFOR
\end{algorithmic}
\end{algorithm}

DtACI does not directly follow the generalized update rule: its time-varying
expert weights introduce additional terms, so the realized error is not the
increment of one scalar coverage state. We instead adapt O2CP at the expert
level.

Algorithm~\ref{alg:dtaci} maintains \(k\) ACI instances, with instance \(i\)
using step size \(\gamma_i\). At time \(t\), an instance \(i_t\) is drawn with
probability \(p_i^{(t)}\). The selected expert constructs its admissible set,
and cross-horizon optimization chooses the control-variable value used for the
final prediction set. Its update uses the error of that final set. Each
unselected expert uses its nominal control-variable value and nominal error, as
in DtACI.

This construction represents the adapted procedure as a time-varying mixture
of ACI experts using the admissible-set construction above.
Lemma~\ref{lemma: boundness of alpha} applies to each expert separately, giving
\(-\gamma_i\leq\alpha_i^{(t)}\leq1+\gamma_i\) for all \(t\). The remaining
argument is the original DtACI weight-change analysis~\cite{gibbs2024aci},
reproduced below.

\begin{proof}
Let
\[
\Phi_t := \sum_i \frac{p_i^{(t)} \alpha_i^{(t)}}{\gamma_i}
\]
and observe that
\begin{align*}
\Phi_t &= \sum_i \frac{p_i^{(t)} (\alpha_i^{(t+1)} - \gamma_i(\alpha - \mathrm{err}_i^{(t)}))}{\gamma_i} \\
&= \sum_i \frac{p_i^{(t)} \alpha_i^{(t+1)}}{\gamma_i} + \sum_i p_i^{(t)} (\mathrm{err}_i^{(t)} - \alpha) \\
&= \Phi_{t+1} + \sum_i \frac{(p_i^{(t)} - p_i^{(t+1)}) \alpha_i^{(t+1)}}{\gamma_i} + \sum_i p_i^{(t)} (\mathrm{err}_i^{(t)} - \alpha).
\end{align*}

Since instance \(i\) is selected with probability \(p_i^{(t)}\),
\(\mathbb{E}[\mathrm{err}_t]=\sum_i p_i^{(t)}\mathrm{err}_i^{(t)}\), which gives
\begin{equation}\label{eq:err_t_alpha}
\mathbb{E}[\mathrm{err}_t] - \alpha = \Phi_t - \Phi_{t+1} + \sum_i \frac{(p_i^{(t+1)} - p_i^{(t)})\alpha_i^{(t+1)}}{\gamma_i}.
\end{equation}

For ease of notation, let $W_t := \sum_i w_i^{(t)}$ and
\[
\tilde{p}_i^{(t+1)} := \frac{p_i^{(t)} \exp(-\eta_t \ell(\beta_t, \alpha_i^{(t)}))}{\sum_{i'} p_{i'}^{(t)} \exp(-\eta_t \ell(\beta_t, \alpha_{i'}^{(t)}))}.
\]
Recall that by definition,
\[
p_i^{(t+1)} = \frac{w_i^{(t+1)}}{\sum_{i'} w_{i'}^{(t+1)}} = (1 - \sigma_t)\tilde{p}_i^{(t+1)} + \frac{\sigma_t}{k}.
\]

A direct calculation gives
\begin{align*}
&\tilde{p}_i^{(t+1)} - p_i^{(t)}\\
&= \frac{p_i^{(t)} \exp(-\eta_t \ell(\beta_t, \alpha_i^{(t)}))}{\sum_{i'} p_{i'}^{(t)} \exp(-\eta_t \ell(\beta_t, \alpha_{i'}^{(t)}))} - p_i^{(t)} \\
&= p_i^{(t)} \frac{\exp(-\eta_t \ell(\beta_t, \alpha_i^{(t)})) - \sum_{i'} p_{i'}^{(t)} \exp(-\eta_t \ell(\beta_t, \alpha_{i'}^{(t)}))}{\sum_{i'} p_{i'}^{(t)} \exp(-\eta_t \ell(\beta_t, \alpha_{i'}^{(t)}))} \\
&= p_i^{(t)} \frac{\sum_{i'} p_{i'}^{(t)} (\exp(-\eta_t \ell(\beta_t, \alpha_i^{(t)})) - \exp(-\eta_t \ell(\beta_t, \alpha_{i'}^{(t)})))}{\sum_{i'} p_{i'}^{(t)} \exp(-\eta_t \ell(\beta_t, \alpha_{i'}^{(t)}))} \\
&= p_i^{(t)} \sum_{i'} \tilde{p}_{i'}^{(t+1)} (\exp(\eta_t \ell(\beta_t, \alpha_{i'}^{(t)}) - \eta_t \ell(\beta_t, \alpha_i^{(t)})) - 1).
\end{align*}

Lemma \ref{lemma: boundness of alpha} gives $\alpha_i^{(t)} \in [-\gamma_i, 1 + \gamma_i]$ and thus $|\ell(\beta_t, \alpha_{i'}^{(t)}) - \ell(\beta_t, \alpha_i^{(t)})| \leq \max\{\alpha, 1 - \alpha\} |\alpha_{i'}^{(t)} - \alpha_i^{(t)}| \leq 1 + 2\gamma_{\max}$. \\
The mean value theorem then gives
\begin{align*}
    &|\exp(\eta_t \ell(\beta_t, \alpha_{i'}^{(t)}) - \eta_t \ell(\beta_t, \alpha_i^{(t)})) - 1|  \\
    &\leq \eta_t(1 + 2\gamma_{\max}) \exp(\eta_t(1 + 2\gamma_{\max})),
\end{align*}

Therefore,
\[
|\tilde{p}_i^{(t+1)} - p_i^{(t)}|
\leq
p_i^{(t)} \eta_t(1+2\gamma_{\max})
\exp(\eta_t(1+2\gamma_{\max})).
\]

Applying Lemma \ref{lemma: boundness of alpha} again, we conclude that
\begin{align*}
& \left| \sum_i
\frac{(p_i^{(t+1)} - p_i^{(t)})\alpha_i^{(t+1)}}{\gamma_i}
\right|\\
&\leq (1-\sigma_t) \sum_i
\left| \frac{(\tilde{p}_i^{(t+1)}-p_i^{(t)})\alpha_i^{(t+1)}}{\gamma_i} \right| \\
&\quad + \sigma_t \sum_i
\left| \frac{(1/k-p_i^{(t)})\alpha_i^{(t+1)}}{\gamma_i} \right| \\
&\leq
\frac{\eta_t(1+2\gamma_{\max})^2}{\gamma_{\min}}
\exp(\eta_t(1+2\gamma_{\max}))
+ 2\sigma_t \frac{1+\gamma_{\max}}{\gamma_{\min}}.
\end{align*}

Summing Equation~\eqref{eq:err_t_alpha} over \(t=1,\ldots,T\) gives
\begin{align*}
\left| \frac{1}{T} \sum_{t=1}^T \mathbb{E}[\mathrm{err}_t] - \alpha \right|
&\leq \frac{|\Phi_1-\Phi_{T+1}|}{T} \\
&\quad + \frac{(1+2\gamma_{\max})^2}{\gamma_{\min}}
\frac{1}{T} \sum_{t=1}^T \eta_t e^{\eta_t(1+2\gamma_{\max})} \\
&\quad + \frac{2(1+\gamma_{\max})}{\gamma_{\min}}
\frac{1}{T} \sum_{t=1}^T \sigma_t.
\end{align*}

Finally, Lemma~\ref{lemma: boundness of alpha} implies
\(\alpha_i^{(t)}/\gamma_i \in [-1,(1+\gamma_i)/\gamma_i]\) for each \(i\).
Therefore
\[
|\Phi_1-\Phi_{T+1}| \leq \frac{1+2\gamma_{\max}}{\gamma_{\min}}.
\]
Substituting this bound into the previous display gives the desired upper
bound.
\end{proof}

Under the learning-rate and mixing schedules of DtACI, the time averages of
the two weight-change terms in the final bound vanish. Hence the adapted
procedure satisfies
\[
\frac1T\sum_{t=1}^T\mathbb E[\mathrm{err}_t]\longrightarrow\alpha.
\]

\section{Experiments}
\subsection{More Details on the Datasets and Base Forecasters}
We evaluate our approach on public real-world datasets spanning autonomous navigation, cyclone trajectory forecasting, and public health, among other domains. The \texttt{Argoverse}\footnote{https://github.com/jagjeet-singh/argoverse-forecasting} autonomous-vehicle dataset contains bird's-eye-view two-dimensional vehicle centroids from urban traffic in Miami and Pittsburgh, sampled at 10~Hz~\cite{Argoverse}. Its training and test sets contain 29,201 and 4,774 sequences, respectively. We adapt the K-NN regressor from the official Argoverse baselines for online prediction. It uses a 10-step observation window and a 5-step prediction horizon and generates 9 predicted trajectories per iteration. We evaluate 250 sequences.

The \cyclone dataset uses forecasts from \texttt{Weather Lab}\footnote{https://deepmind.google.com/science/weatherlab}~\cite{weatherlab3, weatherlab2}. It contains 26 historical cyclone scenarios, each lasting at least four days, with five predicted trajectories over an eight-step horizon. The \flu dataset is constructed from weekly hospitalization records in the CDC FluSight Forecast Hub\footnote{https://github.com/cdcepi/FluSight-forecast-hub}~\cite{mathis2024evaluation}. \weather~\cite{lai2018modeling} and \elec~\cite{xu2023sequential} are standard time-series benchmarks.

\subsection{Standard Deviations for Main Results}
Table~\ref{tab:kdd_main_results_std} reports the standard deviations corresponding to the means in Table~\ref{tab:kdd_main_results}.
\begin{table*}[t]
\centering
{\small
\setlength{\tabcolsep}{3.5pt}%
\begin{tabular}{@{}l|l|ccc|ccc|ccc@{}}
\toprule
Dataset & Metric & ACI+ & DtACI+ & CPID+ & ACI & DtACI & CPID & acMCP & CF-RNN & CopulaCPTS \\
\midrule
\multirow{4}{*}{lane} & SARegret@0.1 & $0.0405$ & $0.0273$ & $0.0510$ & $0.0356$ & $0.0261$ & $0.0419$ & $0.0283$ & $0.0262$ & $0.0350$ \\
 & SARegret@All & $0.0207$ & $0.0183$ & $0.0257$ & $0.0186$ & $0.0165$ & $0.0216$ & $0.0432$ & $0.0400$ & $0.0318$ \\
 & IntervalWidth & $1.177$ & $0.937$ & $0.850$ & $1.345$ & $1.080$ & $1.102$ & $2.540$ & $1.902$ & $0.873$ \\
 & CS & $0.0548$ & $0.0683$ & $0.0686$ & $0.0500$ & $0.0633$ & $0.0616$ & $0.0681$ & $0.0551$ & $0.0882$ \\
\midrule
\multirow{4}{*}{cyclone} & SARegret@0.1 & $0.0089$ & $0.0074$ & $0.0066$ & $0.0103$ & $0.0080$ & $0.0078$ & $0.0136$ & $0.0084$ & $0.0145$ \\
 & SARegret@All & $0.0067$ & $0.0049$ & $0.0064$ & $0.0079$ & $0.0052$ & $0.0062$ & $0.0271$ & $0.0224$ & $0.0344$ \\
 & IntervalWidth & $0.186$ & $0.145$ & $0.143$ & $0.219$ & $0.183$ & $0.191$ & $0.592$ & $0.263$ & $0.321$ \\
 & CS & $0.0259$ & $0.0352$ & $0.0387$ & $0.0263$ & $0.0342$ & $0.0302$ & $0.0448$ & $0.0311$ & $0.0231$ \\
\midrule
\multirow{4}{*}{elec} & SARegret@0.1 & $0.0007$ & $0.0007$ & $0.0002$ & $0.0000$ & $0.0005$ & $0.0000$ & $0.0000$ & $0.0000$ & $0.0000$ \\
 & SARegret@All & $0.0004$ & $0.0005$ & $0.0001$ & $0.0000$ & $0.0005$ & $0.0000$ & $0.0000$ & $0.0000$ & $0.0000$ \\
 & IntervalWidth & $0.0003$ & $0.0008$ & $0.0002$ & $0.0000$ & $0.0004$ & $0.0000$ & $0.0000$ & $0.0000$ & $0.0000$ \\
 & CS & $0.0006$ & $0.0010$ & $0.0003$ & $0.0000$ & $0.0004$ & $0.0000$ & $0.0000$ & $0.0000$ & $0.0000$ \\
\midrule
\multirow{4}{*}{hosp} & SARegret@0.1 & $70$ & $217$ & $46$ & $0$ & $33$ & $0$ & $0$ & $0$ & $0$ \\
 & SARegret@All & $26$ & $117$ & $4$ & $0$ & $101$ & $0$ & $0$ & $0$ & $0$ \\
 & IntervalWidth & $19$ & $125$ & $9$ & $0$ & $102$ & $0$ & $0$ & $0$ & $0$ \\
 & CS & $0.0012$ & $0.0019$ & $0.0003$ & $0.0000$ & $0.0012$ & $0.0000$ & $0.0000$ & $0.0000$ & $0.0000$ \\
\midrule
\multirow{4}{*}{weather} & SARegret@0.1 & $0.036$ & $0.050$ & $0.002$ & $0.000$ & $0.047$ & $0.000$ & $0.000$ & $0.000$ & $0.000$ \\
 & SARegret@All & $0.007$ & $0.027$ & $0.003$ & $0.000$ & $0.046$ & $0.000$ & $0.000$ & $0.000$ & $0.000$ \\
 & IntervalWidth & $0.005$ & $0.043$ & $0.013$ & $0.000$ & $0.007$ & $0.000$ & $0.000$ & $0.000$ & $0.000$ \\
 & CS & $0.0001$ & $0.0011$ & $0.0001$ & $0.0000$ & $0.0013$ & $0.0000$ & $0.0000$ & $0.0000$ & $0.0000$ \\
\bottomrule
\end{tabular}%
}
\caption{Standard deviations across evaluation subsets corresponding to the mean results in Table~\ref{tab:kdd_main_results}.}
\label{tab:kdd_main_results_std}
\end{table*}

\subsection{Uncalibrated Base-Forecaster Performance}
\label{subsec:base_forecaster_calibration}
\vspace{0.5\baselineskip}
We test whether the raw \lane and \cyclone predictive samples are calibrated without conformal correction. At each forecast origin and horizon, we fit a bivariate Gaussian with sample mean $\mu$ and sample covariance $\Sigma$.
\par\vspace{\baselineskip}
\noindent At target coverage $c=1-\alpha$, the prediction set is the ellipse
\[
    \left\{y : (y-\mu)^\top\Sigma^{-1}(y-\mu)
    \leq \chi^2_2(c)\right\},
\]
whose area is $\pi\chi^2_2(c)\sqrt{\det(\Sigma)}$. We use all nine \lane forecast samples and all finite members of the \cyclone \texttt{all\_samples} forecasts.

The evaluation follows the same held-out splits as the main experiments (evaluation configurations 34 and 35): the first 50\% of the 500 \lane cases and the first 66\% of the \cyclone tracks are reserved for pretraining, leaving 250 and 26 evaluation subsets, respectively. We apply each dataset's configured starting time and evaluate the same eleven miscoverage levels used in the main experiments. Coverage and area are first aggregated over time and horizons within each subset, after which subsets are averaged with equal weight.

\begin{table*}[t]
\centering
\small
\setlength{\tabcolsep}{8pt}
\begin{tabular}{@{}lrrrrr@{}}
\toprule
Dataset & Subsets & Coverage@90\% & Area@90\% & CS & Average area \\
\midrule
\lane    & 250 & 69.8\% & 1.665 & 0.190 & 0.975 \\
\cyclone & 26  & 61.1\% & 0.995 & 0.210 & 0.582 \\
\bottomrule
\end{tabular}
\caption{Gaussian fixed-forecaster diagnostic on the held-out \lane and \cyclone subsets. CS and average area are computed over all eleven target coverage levels.}
\label{tab:gaussian_base_forecaster}
\end{table*}

Table~\ref{tab:gaussian_base_forecaster} shows that the nominal 90\% Gaussian sets achieve only 69.8\% coverage on \lane and 61.1\% on \cyclone, under-covering by 20.2 and 28.9 percentage points, respectively. Their calibration scores, 0.190 and 0.210, are also higher (worse) than those of all ACI, DtACI, and CPID variants in Table~\ref{tab:kdd_main_results}, which range from 0.081 to 0.143 on \lane and from 0.065 to 0.099 on \cyclone. Thus, the raw forecast samples do not by themselves provide calibrated uncertainty, whereas the conformal methods materially improve calibration under the shared fixed forecasters. This diagnostic is not a general comparison with parametric, Bayesian, or model-native uncertainty-quantification systems.

Areas are reported in native squared coordinates: squared lane-coordinate units for \lane and squared latitude/longitude degrees for \cyclone. The values are therefore not comparable across datasets. They also cannot be interpreted independently of coverage: the relatively small Gaussian ellipses partly reflect substantial undercoverage rather than greater prediction-set efficiency.

\subsection{Experimental Setup}
\paragraph{Computing infrastructure.}
All experiments run on a Linux system and use the CPU only; no GPU is required at any stage. Our method is computationally lightweight: each dataset--method combination takes approximately 30 seconds to 2 minutes, and every result reported in this paper can be reproduced on a single commodity desktop machine without specialized hardware or unusual memory requirements. The implementation is written in Python and depends only on standard scientific-computing libraries (NumPy, SciPy, pandas, and scikit-learn); the exact pinned version of every dependency is listed in the \texttt{requirements.txt} file distributed with the supplementary code.

\paragraph{Randomness and seeds.}
The \lane and \cyclone experiments are deterministic given the released data, since the base forecasters are fixed and their predictive samples are recorded in advance. For \flu, \weather, and \elec, each evaluation subset is indexed by an integer seed, and both the NumPy and the Python global random number generators are seeded with that value before the subset is run. The seed list for each dataset is stored in the corresponding \texttt{configs/data/} file in the supplementary code, so every reported number can be regenerated exactly by rerunning the released configurations.

\paragraph{Hyperparameters.}
The \texttt{configs/} directory in the supplementary code provides the complete set of final hyperparameters for every method, dataset, and reported experiment, together with the search ranges used during tuning. We tune these parameters on 250 \lane sequences that are disjoint from the evaluation set. A Bayesian optimization framework~\cite{frazier2018tutorial_bayesian_opt} tunes the learning rate ($\eta$), which controls calibration-threshold updates from coverage errors; the number of $\beta$ values in the empirical distribution $F$ ($B$); the learning rate for horizon-wise calibration-threshold updates; and the threshold for selecting $\beta$ values in $F$. We select the configuration minimizing the tuning objective over 100 iterations, which take approximately 5 hours. For the other datasets, we generally use similar hyperparameter settings without further tuning.

\subsection{Evaluation}
Our main evaluation metrics are calibration score~\cite{kuleshov2018accurate}, interval width, and strongly adaptive regret (SARegret)~\cite{daniely2015strongly, Zhangsariodr}. For each time step \(t\), miscoverage rate $\alpha_i$, and horizon $h$, let $\mathrm{cov}_t^h(\alpha_i)$ equal 1 if the prediction set covers the ground truth and 0 otherwise.

\textbf{Interval Width:}
For the one-dimensional datasets \flu, \weather, and \elec, interval width is the distance between the upper and lower bounds. For the two-dimensional \lane and \cyclone datasets, interval width denotes the area of the prediction set.

\textbf{Calibration Score (CS):}
CS is the average absolute difference between target and empirical coverage across a range of miscoverage rates. For a model $M$, let $k_M(c)$ denote the proportion of prediction distributions that contain the observed outcome at confidence level $c = 1-\alpha$. A perfectly calibrated model satisfies $k_M(c) = c$ for all $c$. We define
\[
CS(M) = \int_0^1 |k_M(c)-c|\, dc,
\]
which we approximate by a finite sum over discretized confidence levels. We then average CS across horizons and sequences.

\textbf{Strongly Adaptive Regret (SARegret):}
SARegret evaluates online learning algorithms in nonstationary environments. Unlike fixed-horizon metrics, it considers all contiguous time intervals. Specifically, SARegret is the maximum static regret over all intervals of length \(\tau\):
\begin{equation}
\begin{aligned}
\operatorname{SARegret}(T,\tau)
&= \max_{[s,s+\tau-1]\subseteq[1,T]}
\Bigg(
\sum_{t=s}^{s+\tau-1}\ell_t(w_t) \\
&\qquad
- \min_{w\in\mathcal{W}}
\sum_{t=s}^{s+\tau-1}\ell_t(w)
\Bigg).
\end{aligned}
\end{equation}
A small SARegret indicates that, on any interval of length \(\tau\), the algorithm's cumulative loss is close to that of the best fixed predictor for that interval. In our experiments, we use $\tau = 15$ and take $\ell_t$ to be the pinball loss.

\subsection{An Example of User-Specified Objective Function: Increasing Uncertainty}
Our method can optimize user-specified objectives. For example, temporal dependence often causes uncertainty to accumulate across forecast horizons~\cite{ouyang2018multi, cachay2025elucidated}. We can therefore encourage prediction intervals to expand with the forecast horizon:
\[
J_{inc} = \min\sum_{h=2}^H\mathrm{max}(0, \lambda_Ju_h-u_{h-1}),
\]
where $\lambda_J \in [0, \infty)$ controls the strength of the objective. When $\lambda_J = 0$, $J_{inc}$ has no effect. As $\lambda_J$ increases, the optimization more strongly favors an interval at horizon $h$ that is larger than the interval at horizon $h-1$.

\subsection{Significance Testing}
We use one-sided Wilcoxon signed-rank tests on per-trajectory (per-subset) metric values to assess each ``$+$'' variant against its paired baseline.
Table~\ref{tab:wilcoxon_subset_pairs} reports one-sided $p$-values for the hypothesis that the ``$+$'' method achieves a lower (better) metric than its baseline. Two-sided $p$-values appear in parentheses.
Bold entries indicate $p < 0.05$ (one-sided).

\begin{table*}[t]
\centering
\small
\caption{Wilcoxon signed-rank $p$-values comparing each ``$+$'' method with its paired baseline (lower is better). The main values are one-sided, and {\scriptsize parenthesized} values are two-sided. \textbf{Bold} indicates a one-sided $p < 0.05$.}
\label{tab:wilcoxon_subset_pairs}
\begin{tabular}{@{}l|l|ccc@{}}
\toprule
Dataset & Metric & ACI$^+$ vs.\ ACI & DtACI$^+$ vs.\ DtACI & CPID$^+$ vs.\ CPID \\
\midrule
\multirow{4}{*}{lane} & SARegret@0.1 & $\mathbf{<\!0.0001}\,{\scriptsize(<\!0.0001)}$ & $\mathbf{<\!0.0001}\,{\scriptsize(<\!0.0001)}$ & $\mathbf{<\!0.0001}\,{\scriptsize(<\!0.0001)}$ \\
 & SARegret@All & $\mathbf{<\!0.0001}\,{\scriptsize(<\!0.0001)}$ & $\mathbf{<\!0.0001}\,{\scriptsize(<\!0.0001)}$ & $\mathbf{<\!0.0001}\,{\scriptsize(<\!0.0001)}$ \\
 & IntervalWidth & $\mathbf{<\!0.0001}\,{\scriptsize(<\!0.0001)}$ & $\mathbf{<\!0.0001}\,{\scriptsize(<\!0.0001)}$ & $\mathbf{<\!0.0001}\,{\scriptsize(<\!0.0001)}$ \\
 & CS & $0.7657\,{\scriptsize(0.4686)}$ & $0.9997\,{\scriptsize(0.0006)}$ & $0.9954\,{\scriptsize(0.0092)}$ \\
\midrule
\multirow{4}{*}{cyclone} & SARegret@0.1 & $\mathbf{0.0315}\,{\scriptsize(0.0630)}$ & $\mathbf{0.0055}\,{\scriptsize(0.0110)}$ & $\mathbf{0.0018}\,{\scriptsize(0.0036)}$ \\
 & SARegret@All & $\mathbf{<\!0.0001}\,{\scriptsize(<\!0.0001)}$ & $\mathbf{<\!0.0001}\,{\scriptsize(<\!0.0001)}$ & $\mathbf{<\!0.0001}\,{\scriptsize(<\!0.0001)}$ \\
 & IntervalWidth & $\mathbf{0.0005}\,{\scriptsize(0.0010)}$ & $\mathbf{<\!0.0001}\,{\scriptsize(<\!0.0001)}$ & $0.9924\,{\scriptsize(0.0163)}$ \\
 & CS & $\mathbf{<\!0.0001}\,{\scriptsize(<\!0.0001)}$ & $\mathbf{<\!0.0001}\,{\scriptsize(<\!0.0001)}$ & $\mathbf{0.0279}\,{\scriptsize(0.0559)}$ \\
\bottomrule
\end{tabular}
\end{table*}

\subsection{More on Ablation Study}
\label{subsec:more_ablation}
\begin{figure*}[t]
    \centering
    \includegraphics[width=0.8\linewidth]{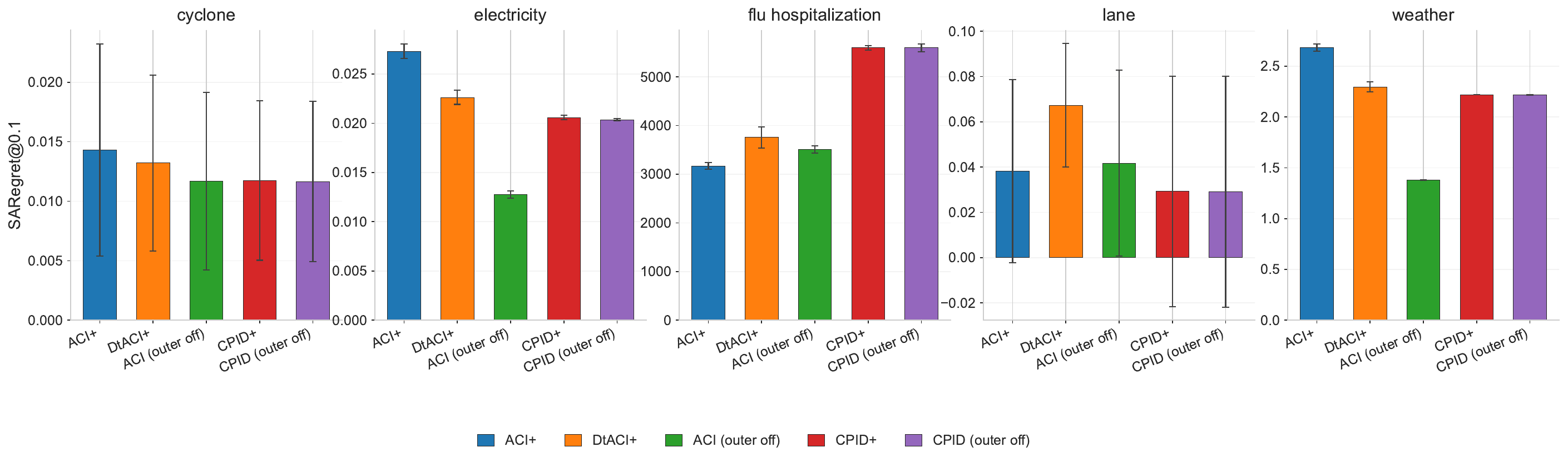}
    \caption{Online CP update ablation: SARegret@0.1. Bars compare full ACI$+$, DtACI$+$, and CPID$+$ to variants with frozen coverage-driven updates. Error bars show standard deviation.}
    \label{fig:ablation_outer_saregret_01}
\end{figure*}

\begin{figure*}[t]
    \centering
    \includegraphics[width=0.8\linewidth]{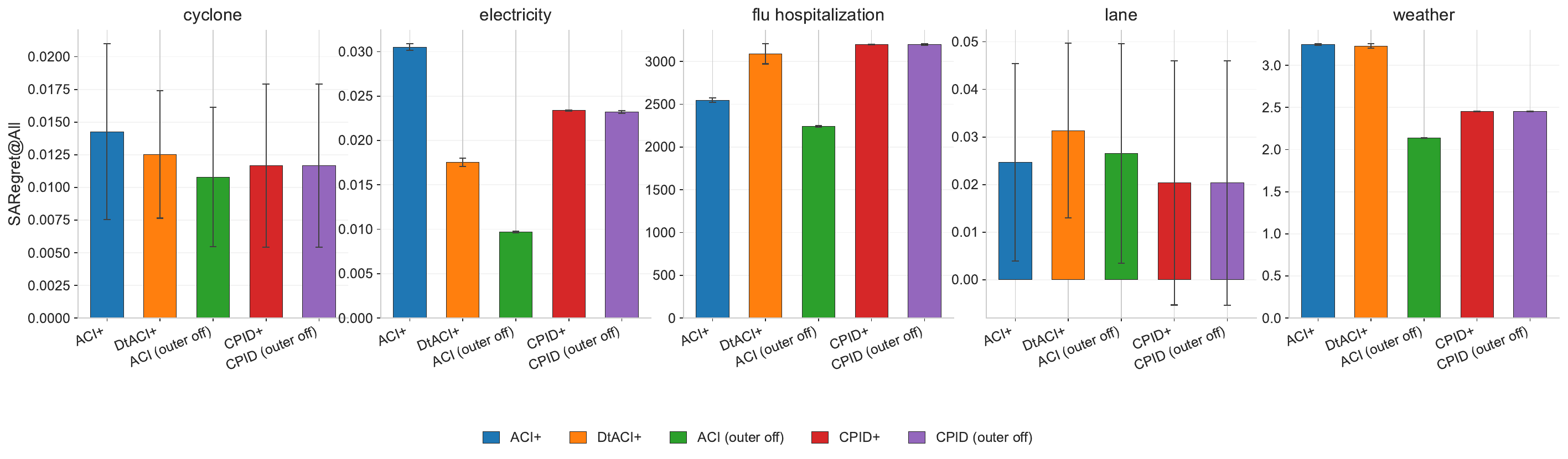}
    \caption{Online CP update ablation: SARegret@All (same legend as Figure~\ref{fig:ablation_outer_saregret_01}).}
    \label{fig:ablation_outer_saregret_all}
\end{figure*}

\begin{figure*}[t]
    \centering
    \includegraphics[width=0.8\linewidth]{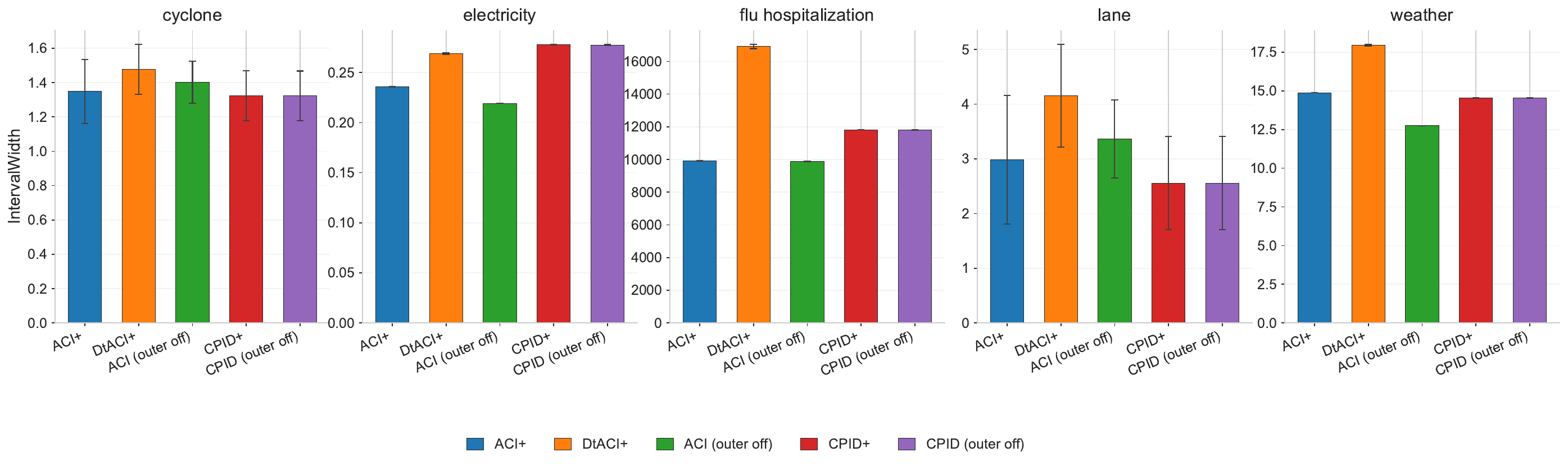}
    \caption{Online CP update ablation: average interval width (area for 2D prediction sets on \lane\ and \cyclone).}
    \label{fig:ablation_outer_width}
\end{figure*}

\begin{figure*}[t]
    \centering
    \includegraphics[width=0.8\linewidth]{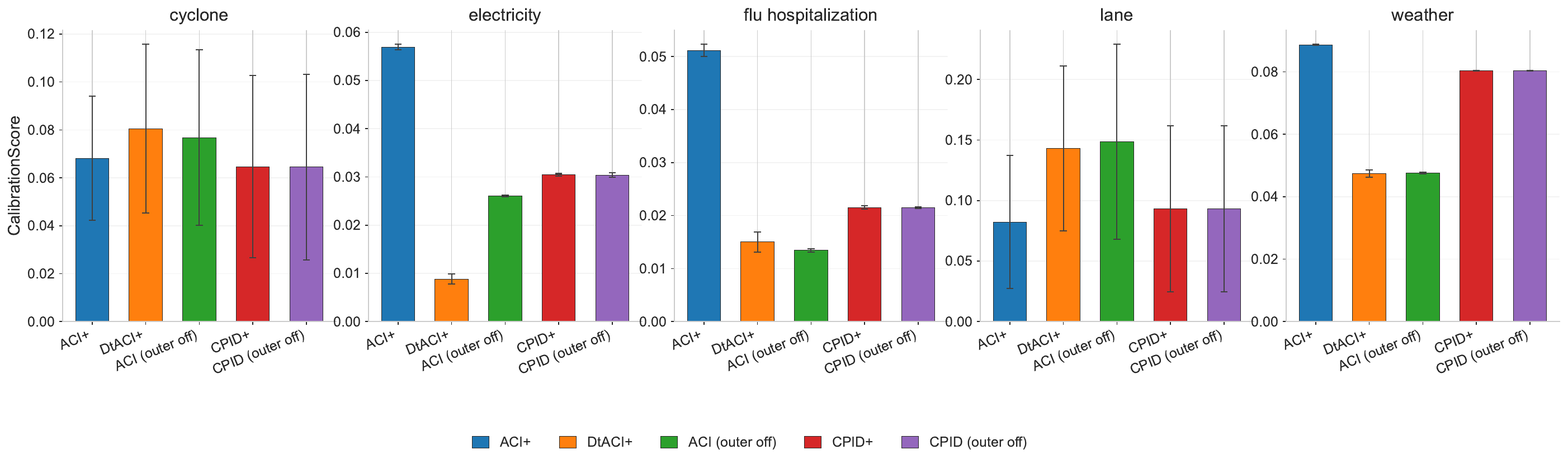}
    \caption{Online CP update ablation: calibration score (lower is better).}
    \label{fig:ablation_outer_cs}
\end{figure*}

As noted in the main paper, comparisons between each ``$+$'' variant and its paired baseline (\aciw/\aci, \dtaciw/\dtaci, \cpidw/\cpid) isolate the contribution of cross-horizon optimization. The baselines fix the control variables at the midpoint of the admissible interval, whereas the ``$+$'' variants optimize over that interval.
To examine the role of the online CP updates, we conduct an additional ablation that freezes the coverage-driven updates while retaining cross-horizon optimization.
For \aciw and \dtaciw, freezing the online CP update means that $\alpha_t$ is no longer adjusted from observed coverage errors. For \cpidw, the quantile $q_t$ is no longer updated from coverage feedback.
Because \aci and \dtaci both update $\alpha_t$, a single experimental configuration suffices as the frozen-update baseline for \emph{both} \aciw and \dtaciw; the plots currently label this bar as ``ACI (outer off),'' and it serves as the common reference when comparing against \dtaciw as well.
Figures~\ref{fig:ablation_outer_saregret_01}--\ref{fig:ablation_outer_cs} show that the frozen-update variants track the full methods closely in SARegret, interval width, and calibration score across all datasets.
These results support the empirical effectiveness of cross-horizon optimization even when the coverage-driven online CP updates are disabled.
Nevertheless, the coverage-driven online CP updates and admissible constraints jointly provide the long-term marginal coverage guarantee in our theoretical construction. Both therefore remain essential components of \ourmethod{} even when their scalar metrics are similar.

\subsection{Interval Size}
\begin{figure*}[h!]
    \centering
    \includegraphics[width=0.95\linewidth]{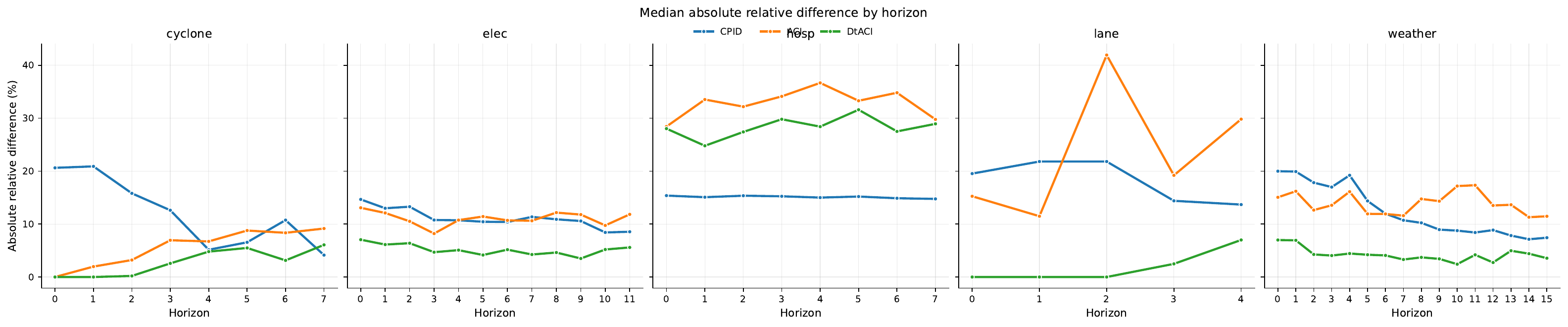}
    \caption{Median absolute relative difference in prediction-interval width
between each ``$+$'' variant and its paired baseline (\aciw/\aci,
\dtaciw/\dtaci, \cpidw/\cpid) at miscoverage rate $\alpha=0.1$, as a function
of forecast horizon. Each curve shows the median across trajectories; higher
values indicate larger interval-width adjustments introduced by the
cross-horizon optimization.}
    \label{fig:horizon_abs_rel_diff_median}
\end{figure*}

We also examine how much our method changes interval width relative to the baseline.
For each trajectory, we compute the absolute relative difference $|s_+ - s| / |s|$, where $s$ and $s_+$ are the interval widths of the baseline and its ``$+$'' variant, respectively.
Figure~\ref{fig:horizon_abs_rel_diff_median} reports the median of this quantity across trajectories at miscoverage rate $\alpha=0.1$ as a function of forecast horizon. The median absolute relative difference remains below approximately 35\% for most dataset--method--horizon combinations, although \aci on \lane peaks at approximately 42\%. The horizon-wise patterns depend on both the dataset and underlying conformal method. On \cyclone, the \cpid difference generally decreases with horizon, whereas the \aci and \dtaci differences increase from values near zero. The curves are comparatively stable on \elec and \flu, more variable on \lane, and generally decline or remain moderate on \weather. Overall, cross-horizon optimization changes interval widths meaningfully without uniformly expanding or shrinking them across forecast horizons.

\end{document}